\documentclass[11pt]{article}

\usepackage[final]{acl}

\usepackage{times}
\usepackage{latexsym}
\usepackage{kotex}
\usepackage{CJKutf8}

\usepackage{xcolor}
\usepackage{booktabs}
\usepackage{amsmath}
\usepackage{amssymb}
\usepackage{subcaption}
\usepackage{pgfplots}
\pgfplotsset{compat=1.17}
\usepgfplotslibrary{groupplots}
\pgfplotsset{compat=1.17}
\pgfplotsset{compat=1.18}
\usepgfplotslibrary{statistics}
\usepackage{multirow}
\usepackage[most]{tcolorbox}
\usepackage{tabularx}
\usepackage{array}
\usepackage{longtable}
\usepackage{graphicx}

\usepackage{tikz}
\usepgfplotslibrary{groupplots}

\newcommand{\kp}[2]{#1 \scorepos{#2}} 
\newcommand{\kn}[2]{#1 \scoreneg{#2}} 
\newcommand{\kz}[1]{#1 \textcolor{gray}{\scriptsize{(+0)}}}

\newcommand{\vp}[2]{#1 \scorepos{#2}}
\newcommand{\vn}[2]{#1 \scoreneg{#2}}
\newcommand{\vz}[1]{#1 \scorezero}

\newcommand{\scorepos}[1]{\textcolor{red}{\scriptsize{(+#1)}}}
\newcommand{\scoreneg}[1]{\textcolor{blue}{\scriptsize{(-#1)}}}
\newcommand{\scorezero}{\textcolor{gray}{\scriptsize{(+0.0)}}}

\usepackage{fvextra}

\newtcolorbox{englishpromptbox}[1]{
  enhanced jigsaw,
  breakable,
  width=\textwidth,
  colback=red!3,
  colframe=red!70!black,
  coltitle=white,
  colbacktitle=red!70!black,
  title=\textbf{#1},
  fonttitle=\bfseries,
  boxrule=0.8pt,
  arc=2mm,
  outer arc=2mm,
  left=3mm,
  right=3mm,
  top=2mm,
  bottom=2mm,
  pad at break*=2mm,
  before skip=6pt,
  after skip=8pt
}

\newtcolorbox{koreanpromptbox}[1]{
  enhanced jigsaw,
  breakable,
  width=\textwidth,
  colback=blue!3,
  colframe=blue!70!black,
  coltitle=white,
  colbacktitle=blue!70!black,
  title=\textbf{#1},
  fonttitle=\bfseries,
  boxrule=0.8pt,
  arc=2mm,
  outer arc=2mm,
  left=3mm,
  right=3mm,
  top=2mm,
  bottom=2mm,
  pad at break*=2mm,
  before skip=6pt,
  after skip=8pt
}

\usepackage[T1]{fontenc}
\usepackage[utf8]{inputenc}

\usepackage{microtype}

\usepackage{inconsolata}

\usepackage{graphicx}

\title{Same Judgment, Different Scores: Numerical Scoring Topology in LLM-as-a-Judge Evaluation}

\author{
 \textbf{Doyun Kim},
 \textbf{Chanwoo Kim},
 \textbf{Jiwoo Kum},
 \textbf{Minsoo Song},
 \textbf{Chanjun Park\textsuperscript{\textdagger}}
\\
\\
 Soongsil University
\\
 \texttt{\{doyunkim, kcw4776, goldi1204, ecoses042\}@soongsil.ac.kr} 
 \\
 \texttt{chanjun.park@ssu.ac.kr}
\\
}

\begin{document}
\maketitle

\begingroup
\renewcommand{\thefootnote}{\textdagger}
\footnotetext{Corresponding authors.}
\endgroup

\begin{abstract}
LLMs are increasingly used as automated evaluators to reduce the cost and time required for human annotation. However, the role of scoring-scheme design in the reliability of LLM-as-a-Judge evaluation remains underexplored. To address this problem, we introduce \textit{numerical scoring topology}, which characterizes the structural design of a scoring scheme in terms of its scale width, granularity, offset, symmetry, and boundary placement. Building on this definition, we conduct the first systematic study of how numerical scoring topology influences zero-shot LLM-as-a-Judge evaluation across diverse benchmarks and judge models. Our results show that numerical scoring topology systematically affects judge scores, calibration, and robustness, particularly in judgment-intensive evaluations. Shifted and narrow ranges tend to induce some of the largest score deviations across many settings. We further find that stronger and larger judge models generally achieve higher discriminative validity and greater robustness to scoring-topology changes. These findings suggest that numerical scoring topology should be treated as an integral component of LLM-as-a-Judge evaluation and highlight the need for more principled and standardized scoring protocols.

% LLMs are increasingly used as automated judges, yet the impact of numerical scoring topology on evaluation reliability remains largely unexplored. We present the first systematic study of how numerical scoring topology influences zero-shot LLM-as-a-Judge evaluation across thirteen scoring topologies, six representative benchmarks, and seven judge models. Our analysis yields three main findings. First, the impact of scoring topology is strongly task dependent: objective reasoning benchmarks remain largely stable across scoring schemes, whereas subjective evaluation tasks exhibit substantially greater sensitivity. Second, controlled ablation studies reveal that numerical boundary alignment is more strongly associated with evaluation instability than score width alone. In particular, small changes from round to nearby non-round score boundaries can increase disagreement rates from 16.0\% to 81.5\%. Third, these effects become consistently more pronounced in cross-lingual evaluation, where disagreement rates increase by up to 28 percentage points compared with monolingual settings. Our findings identify numerical scoring topology as an important yet previously overlooked factor in LLM-as-a-Judge evaluation and provide empirical evidence for developing more reliable and reproducible automated evaluation protocols.
\end{abstract}

\begin{figure}[t]
    \centering
    \includegraphics[
        width=\columnwidth
    ]{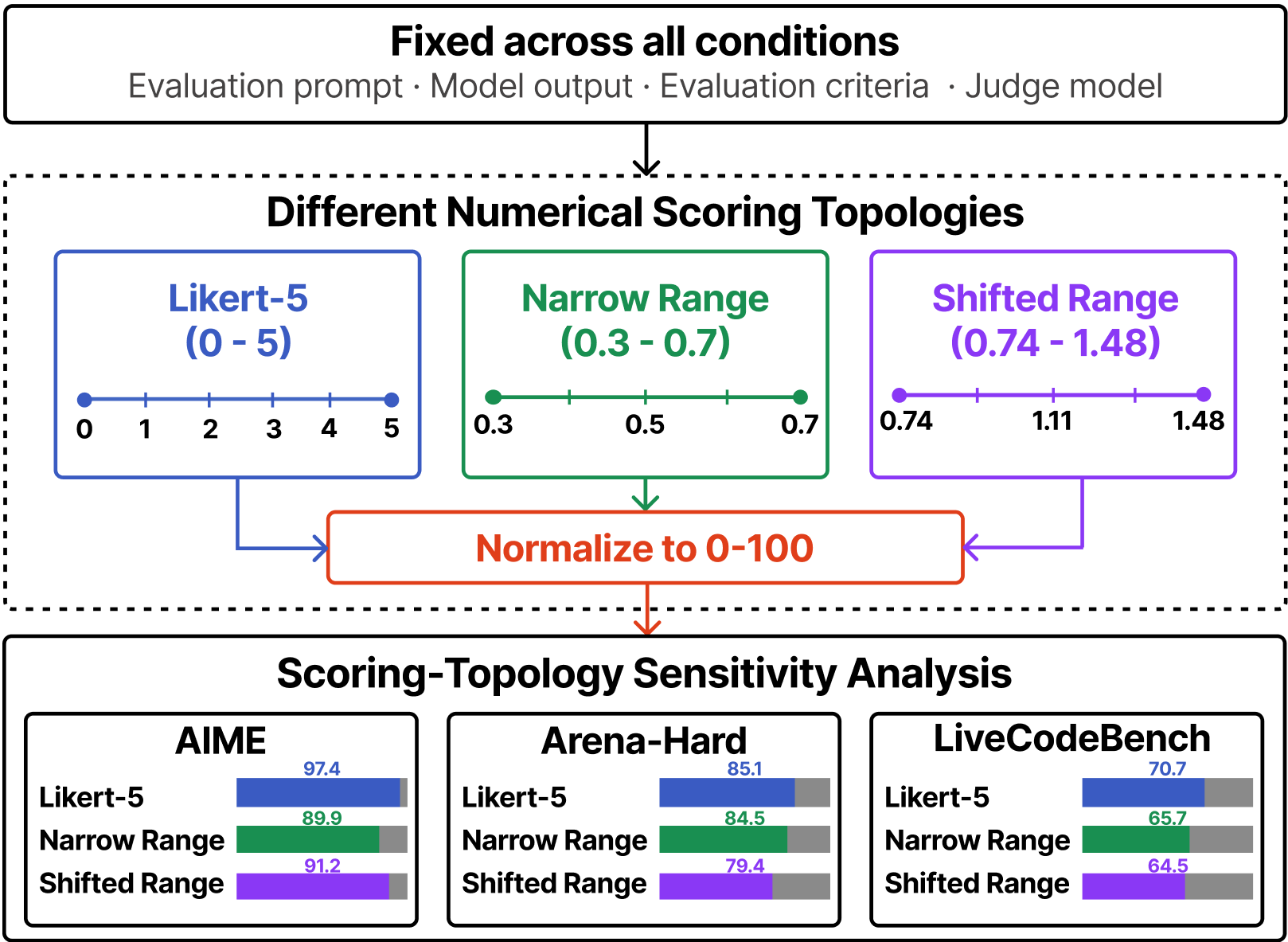}
    \caption{
    Overview of the scoring-topology sensitivity analysis.
    We keep the evaluation prompt, model output, evaluation criteria, and judge model fixed while varying only the numerical scoring topology.
    Scores are normalized and compared across representative benchmarks.
    }
    \label{fig:scoring_topology_overview}
\end{figure}

\section{Introduction}
\label{sec:introduction}
Large Language Models (LLMs) are increasingly used as automated evaluators, a paradigm commonly known as LLM-as-a-Judge~\cite{zheng2023judging, liu2023g}. By reducing the need for costly and time consuming human annotation, LLM judges provide a scalable approach to evaluating model outputs across diverse tasks, including mathematical, reasoning, code generation, instruction following, and dialogue~\cite{tan2024judgebench, tong2024codejudge}. As automated evaluation plays an increasingly central role in model development and benchmarking, ensuring the reliability and reproducibility of LLM judges has become an important research challenge.

% Large Language Models (LLMs) are increasingly used as automated evaluators, a paradigm commonly known as \textit{LLM-as-a-Judge}. By replacing expensive and time-consuming human annotation with model-based evaluation, LLM judges have become a standard component of modern NLP research, supporting benchmarks for mathematical reasoning, code generation, instruction following, dialogue systems, and many other generation tasks. As automated evaluation continues to play a central role in model development and benchmarking, ensuring the reliability and reproducibility of LLM judges has become an increasingly important research challenge.

Prior studies have identified factors that influence the reliability of LLM judges, including candidate response order, response length, model identity, prompt formulation, and differences in evaluator calibration~\cite{zheng2023judging, zhuo2024prosa, chen2025beyond}. To mitigate these sources of variability, researchers have explored improved evaluation prompts, explicit scoring criteria, model alignment, and judge models specifically trained for evaluation~\cite{liu2025judge, kim2024prometheus}. Most of these studies concentrate on the semantic and behavioral aspects of the evaluation process.

% Prior studies have shown that the reliability of LLM judges is influenced by various factors, including the order of candidate responses, response length, model identity, prompt formulation, and differences in evaluator calibration. In response, existing work has explored methods such as improved evaluation prompts, explicit scoring criteria, model alignment, and judge models trained specifically for evaluation. Most of these studies concentrate on the semantic and behavioral aspects of the evaluation process.

% A growing body of work has shown that LLM judges are affected by numerous sources of bias. Previous studies have identified position bias, verbosity bias, self-preference bias, prompt sensitivity, evaluator calibration issues, and various architectural limitations. Consequently, most recent research has focused on improving evaluation reliability through better prompting strategies, alignment methods, specialized evaluator models, and refined evaluation criteria.

However, the numerical scoring scheme itself remains comparatively underexplored. Existing evaluation protocols employ diverse scoring formats, including binary decisions, Likert scales, percentage scores, bounded integer ranges, and custom numerical intervals~\cite{li2025generation, lee2024checkeval}. Researchers often select these formats based on convenience, benchmark compatibility, or downstream evaluation requirements. Despite their structural differences, these scoring schemes are commonly treated as equivalent representations of the same underlying judgment and are assumed to produce comparable evaluator behavior after normalization.

% However, the numerical scoring scheme itself remains comparatively underexplored. Existing evaluation protocols use a wide range of scoring formats, including binary decisions, Likert scales, percentage scores, bounded integer ranges, and custom numerical intervals. Researchers typically select these formats based on convenience, compatibility with existing benchmarks, or downstream evaluation requirements. Despite their structural differences, these scoring schemes are generally treated as equivalent representations of the same underlying judgment. Different numerical scales are therefore assumed to produce comparable evaluator behavior after normalization.

% In contrast, one component of LLM evaluation has received surprisingly little attention: the numerical scoring scheme itself. Existing evaluation protocols employ a wide variety of scoring formats, including binary decisions, Likert scales, percentage scores, bounded integer ranges, and customized numerical intervals. These scoring schemes are typically selected for convenience, compatibility with existing benchmarks, or downstream evaluation requirements. More importantly, they are almost universally treated as interchangeable representations of the same underlying judgment. After simple normalization, different numerical scales are generally assumed to preserve identical evaluator behavior.

In this paper, we demonstrate that this assumption does not consistently hold. As illustrated in Figure~\ref{fig:scoring_topology_overview}, we keep the evaluation prompt, model output, evaluation criteria, and judge model fixed while varying only the numerical scoring topology. This change alone substantially affects the judgments produced by zero-shot LLM judges. Small changes to score design can lead to substantial differences in score distributions and calibration patterns. The magnitude of this effect varies across tasks: some evaluations show limited sensitivity, whereas others produce less consistent outcomes under semantically equivalent evaluation settings.

% In this paper, we show that this assumption does not always hold. Even when evaluation prompts, model outputs, and evaluation criteria remain identical, changing only the numerical topology of the scoring space can substantially alter the behavior of zero-shot LLM judges. Seemingly minor modifications to score boundaries, numerical offsets, or boundary alignment can produce noticeably different score distributions and calibration behavior. While some evaluation tasks remain largely unaffected, others become highly sensitive to numerical scoring topology, resulting in inconsistent evaluation outcomes despite identical semantic evaluation conditions.

To investigate this phenomenon systematically, we introduce \textit{numerical scoring topology}, which we use operationally to denote the structural configuration of a numerical score space rather than topology in the formal mathematical sense. Specifically, it characterizes a scoring scheme in terms of its scale width, granularity, offset, symmetry, and boundary placement. We conduct an empirical study across thirteen scoring topologies and diverse evaluation benchmarks covering both answer-verification and judgment-intensive tasks. Our experimental design varies only the numerical scoring topology while keeping the evaluation prompts, criteria, model outputs, and judge models fixed.

% To investigate this phenomenon systematically, we introduce \textit{numerical scoring topology}, which characterizes the structural design of a scoring scheme in terms of its scale width, granularity, offset, symmetry, and boundary placement. We conduct an empirical study across thirteen scoring topologies, diverse evaluation benchmarks covering both answer-verification and judgment-intensive tasks. Our experimental design varies only the numerical scoring topology while keeping the evaluation prompts, criteria, model outputs, and judge models fixed. This allows us to isolate the effect of numerical score design from other factors.

% To systematically investigate this phenomenon, we define the notion of \textit{numerical scoring topology}, which refers to the structural design of a scoring scheme, including its score boundaries, numerical range, granularity, offsets, and boundary alignment. We then conduct a large-scale empirical study spanning thirteen representative scoring topologies, six evaluation benchmarks covering both objective and subjective tasks, and seven LLM judge models. By varying only the numerical scoring topology while holding prompts, evaluation criteria, and judge models constant, our experimental design isolates the effect of numerical score design from other confounding factors. We further extend our analysis to cross-lingual evaluation to examine whether multilingual settings amplify this sensitivity.

Our experiments show that the effect of numerical scoring topology on LLM judge behavior varies across evaluation tasks. With the Shifted Range (0.74–1.48), the normalized score drops from 97.4 to 89.9 on Arena-Hard and from 70.7 to 65.7 on AIME. A similar decline appears on LiveCodeBench, where the score decreases from 85.1 with Likert-5 to 79.4 with the Narrow Range (0.3–0.7). Overall, these results indicate that numerical scoring schemes are not interchangeable, even after normalization, and that score design can materially affect LLM-based evaluation outcomes.

Our study addresses the following research questions:

% Our study is organized around the following research questions:

\begin{itemize}
    \item \textbf{RQ1: How does numerical scoring topology influence LLM-as-a-Judge evaluation across different scoring schemes and evaluation tasks?} 

    % We first examine whether changing only the numerical scoring topology alters evaluation outcomes. We then conduct controlled ablation studies to identify the effects of score width, boundary placement, and boundary alignment, and analyze whether tasks based primarily on verifiable correctness and tasks requiring greater evaluator discretion exhibit different levels of sensitivity.

    \item \textbf{RQ2: Which properties of numerical scoring topology drive evaluation instability?} 

    % We compare matched and cross-lingual evaluation settings, including English responses evaluated in English, English responses evaluated in Korean, Korean responses evaluated in English, and Korean responses evaluated in Korean, to determine how response language, evaluation language, and language mismatch influence evaluation stability.

    \item \textbf{RQ3: How does the interaction between response language and evaluation language affect sensitivity to numerical scoring topology?}

    \item \textbf{RQ4: How consistently do the effects of numerical scoring topology generalize across different response generators and LLM judge models?} 

    \item \textbf{RQ5: How effectively can prompt-based strategies mitigate evaluation instability induced by numerical scoring topology?}
    
    % We generate responses and conduct evaluations using diverse model families to examine whether the observed effects persist across generator and judge models, and whether their magnitude varies across different generator--judge combinations.
\end{itemize}

\section{Related Work}

\paragraph{LLM-as-a-Judge.}
LLMs increasingly serve as automated evaluators across a broad range of natural language processing (NLP) tasks, reducing reliance on costly human annotation~\cite{zheng2023judging, liu2023g}. Prior studies demonstrate substantial agreement between LLM judgments and human preferences across dialogue evaluation, summarization, instruction following, reasoning, and code generation~\cite{tong2024codejudge, chen2025beyond}. LLM judges commonly evaluate responses through pairwise preference judgment or pointwise numerical scoring based on predefined criteria~\cite{kim2024prometheus}.

% Large Language Models (LLMs) have recently emerged as reliable automatic evaluators for a wide range of natural language generation tasks, substantially reducing the need for expensive human annotation. Recent studies have demonstrated strong agreement between LLM judges and human preferences across diverse benchmarks, including dialogue evaluation, summarization, instruction following, reasoning, and code generation~\cite{}. Consequently, LLM-as-a-Judge has become a standard evaluation paradigm in modern NLP, where judge models are commonly asked to produce either pairwise preferences or numerical quality scores under predefined evaluation criteria~\cite{}.

Efforts to improve LLM judge reliability focus on prompt design, explicit evaluation criteria, calibration, model alignment, and judge-specific training~\cite{liu2025judge, kim2024prometheus}. These approaches mainly address the semantic and behavioral components of evaluation, while the structural design of the numerical scoring space rarely serves as an independent experimental variable.

\paragraph{Numerical Scoring Design in LLM Evaluation.}
Numerical scores are widely used in automatic evaluation, with common schemes including binary labels, Likert scales, percentage scores, bounded integer ranges, and custom intervals~\cite{kim2024prometheus, lee2024checkeval, li2025generation}. These schemes are typically selected according to benchmark conventions, implementation preferences, or downstream requirements. Despite their structural differences, they are commonly treated as equivalent after normalization.

However, this assumption remains largely unexamined. The structural design of the numerical scoring space, including score boundaries, range width, numerical offsets, and boundary alignment, rarely serves as an independent experimental variable. Our work addresses this problem by treating numerical score design as an independent factor in LLM-as-a-Judge evaluation.

\begin{table*}[t]
\centering
\scriptsize
\setlength{\tabcolsep}{4pt}
\resizebox{\textwidth}{!}{%
\begin{tabular}{lcccccc}
\toprule
Scoring Scale & MATH-500 & AIME & GPQA Main & LCB & Arena-Hard & IFEval \\
\midrule
Unit-interval (0 to 1) & \vn{85.7}{0.8} & \vn{67.4}{3.3} & \vn{73.1}{4.4} & \vn{84.6}{0.5} & \vn{94.9}{2.5} & \vn{80.2}{3.0} \\
\textbf{Likert-5 (0 to 5)} & \textbf{\vz{86.5}} & \textbf{\vz{70.7}} & \textbf{\vz{77.5}} & \textbf{\vz{85.1}} & \textbf{\vz{97.4}} & \textbf{\vz{83.2}} \\
Likert-10 (0 to 10) & \vp{86.8}{0.3} & \vn{69.7}{1.0} & \vp{77.8}{0.3} & \vn{82.9}{2.2} & \vn{93.3}{4.1} & \vn{81.4}{1.8} \\
Linear-50 (0 to 50) & \vp{87.2}{0.7} & \vp{71.3}{0.6} & \vp{78.6}{1.1} & \vn{84.3}{0.8} & \vn{94.8}{2.6} & \vn{82.5}{0.7} \\
Linear-100 (0 to 100) & \vp{87.8}{1.3} & \vp{71.0}{0.3} & \vp{78.0}{0.5} & \vn{84.1}{1.0} & \vn{93.3}{4.1} & \vn{82.5}{0.7} \\
Linear-1000 (0 to 1000) & \vp{87.4}{0.9} & \vn{70.2}{0.5} & \vn{77.3}{0.2} & \vn{81.3}{3.8} & \vn{93.9}{3.5} & \vn{80.8}{2.4} \\
Positive Offset (3 to 15) & \vn{85.8}{0.7} & \vn{68.3}{2.4} & \vn{76.7}{0.8} & \vn{83.3}{1.8} & \vn{94.3}{3.1} & \vn{81.4}{1.8} \\
Narrow Range (0.3 to 0.7) & \vn{85.2}{1.3} & \vn{67.8}{2.9} & \vn{73.8}{3.7} & \textbf{\vn{79.4}{5.7}} & \vn{88.3}{9.1} & \textbf{\vn{77.2}{6.0}} \\
Symmetric Range ($-5$ to 5) & \vp{87.2}{0.7} & \vn{68.9}{1.8} & \vn{75.9}{1.6} & \vn{84.2}{0.9} & \vp{97.6}{0.2} & \vn{82.0}{1.2} \\
Asymmetric Range ($-5$ to 2) & \vp{88.2}{1.7} & \vn{68.7}{2.0} & \vp{80.0}{2.5} & \vn{83.3}{1.8} & \vp{98.0}{0.6} & \vn{82.7}{0.5} \\
Shifted Range (0.74 to 1.48) & \vn{85.4}{1.1} & \textbf{\vn{65.7}{5.0}} & \textbf{\vn{73.0}{4.5}} & \vn{80.8}{4.3} & \textbf{\vn{89.9}{7.5}} & \vn{77.6}{5.6} \\
Irregular Range I ($-3.7$ to 8.9) & \textbf{\vp{89.1}{2.6}} & \vp{71.1}{0.4} & \vp{78.2}{0.7} & \vn{84.1}{1.0} & \vn{94.9}{2.5} & \vn{82.3}{0.9} \\
Irregular Range II ($-2.35$ to 7.85) & \vp{88.3}{1.8} & \vp{71.9}{1.2} & \vp{78.1}{0.6} & \vn{84.7}{0.4} & \vn{94.5}{2.9} & \vp{83.3}{0.1} \\
\bottomrule
\end{tabular}%
}

\caption{\textbf{Mean normalized judge scores across diverse scoring topologies.} GPT-5-mini is used both to generate the benchmark responses and to evaluate them under all scoring topologies. All raw scores are normalized to a 0--100 scale for comparison. Values in parentheses denote the deviation relative to the \textbf{Likert-5} baseline; red indicates an increase and blue indicates a decrease in normalized judge scores.}

\label{tab:scoring-scale-results}
\end{table*}

\begin{table*}[t]
\centering
\small
\begin{tabular*}{\textwidth}{@{\extracolsep{\fill}}lcccccc@{}}
\toprule
Scoring Scale & MATH-500 & AIME & GPQA Main & LCB & Arena-Hard & IFEval \\
\midrule
\multicolumn{7}{c}{Disagreement Rate ($|\Delta| > 10$)} \\
\midrule
Unit-interval (0 to 1) & 9.0 (45) & 20.8 (25) & 20.8 (93) & 9.7 (17) & 4.8 (24) & 28.1 (152) \\
Likert-10 (0 to 10) & 4.8 (24) & 10.0 (12) & 6.9 (31) & 8.0 (14) & 6.2 (31) & 20.0 (108) \\
Linear-50 (0 to 50) & 4.8 (24) & 6.7 (8) & 7.6 (34) & 8.6 (15) & 6.4 (32) & 21.4 (116) \\
Linear-100 (0 to 100) & 5.0 (25) & 9.2 (11) & 7.6 (34) & 5.7 (10) & 6.6 (33) & 24.2 (131) \\
Linear-1000 (0 to 1000) & 5.8 (29) & 8.3 (10) & 7.1 (32) & 10.3 (18) & 6.2 (31) & 20.7 (112) \\
Positive Offset (3 to 15) & 8.4 (42) & 17.5 (21) & 17.0 (76) & 12.6 (22) & 9.4 (47) & 27.7 (150) \\
Narrow Range (0.3 to 0.7) & 9.0 (45) & 9.2 (11) & 13.6 (61) & 19.4 (34) & 23.4 (117) & 32.2 (174) \\
Symmetric Range ($-5$ to 5) & 5.2 (26) & 18.3 (22) & 11.4 (51) & 9.7 (17) & 5.0 (25) & 21.4 (116) \\
Asymmetric Range ($-5$ to 2) & 8.6 (43) & 16.7 (20) & 12.1 (54) & 13.1 (23) & 7.2 (36) & 25.7 (139) \\
Shifted Range (0.74 to 1.48) & 8.8 (44) & \textbf{23.3} (28) & \textbf{22.5} (101) & \textbf{19.4} (34) & \textbf{28.8} (144) & \textbf{36.4} (197) \\
Irregular Range I ($-3.7$ to 8.9) & 8.8 (44) & 18.3 (22) & 15.8 (71) & 12.0 (21) & 7.6 (38) & 25.5 (138) \\
Irregular Range II ($-2.35$ to 7.85) & \textbf{9.6} (48) & 16.7 (20) & 18.3 (82) & 13.1 (23) & 7.6 (38) & 28.8 (156) \\
\midrule
\multicolumn{7}{c}{Disagreement Rate ($|\Delta| > 20$)} \\
\midrule
Unit-interval (0 to 1) & 2.6 (13) & 3.3 (4) & 2.9 (13) & 5.7 (10) & 1.8 (9) & 15.2 (82) \\
Likert-10 (0 to 10) & 1.4 (7) & 1.7 (2) & 1.6 (7) & 4.6 (8) & 1.0 (5) & 11.3 (61) \\
Linear-50 (0 to 50) & 1.2 (6) & 0.8 (1) & 2.0 (9) & 4.6 (8) & 1.8 (9) & 12.4 (67) \\
Linear-100 (0 to 100) & 2.2 (11) & 3.3 (4) & 1.1 (5) & 2.9 (5) & 1.0 (5) & 12.6 (68) \\
Linear-1000 (0 to 1000) & 2.8 (14) & 1.7 (2) & 1.8 (8) & 6.3 (11) & 2.8 (14) & 12.4 (67) \\
Positive Offset (3 to 15) & 2.4 (12) & 2.5 (3) & 3.8 (17) & 5.1 (9) & 3.4 (17) & 15.9 (86) \\
Narrow Range (0.3 to 0.7) & 1.8 (9) & 1.7 (2) & 2.0 (9) & \textbf{8.6} (15) & \textbf{7.8} (39) & \textbf{18.1} (98) \\
Symmetric Range ($-5$ to 5) & 2.8 (14) & 3.3 (4) & 1.8 (8) & 6.3 (11) & 1.6 (8) & 12.4 (67) \\
Asymmetric Range ($-5$ to 2) & 4.2 (21) & 2.5 (3) & \textbf{5.4} (24) & 4.6 (8) & 2.6 (13) & 13.5 (73) \\
Shifted Range (0.74 to 1.48) & 2.4 (12) & 5.0 (6) & 2.9 (13) & 6.9 (12) & 6.4 (32) & 13.7 (74) \\
Irregular Range I ($-3.7$ to 8.9) & 4.4 (22) & 2.5 (3) & 4.9 (22) & 5.1 (9) & 3.0 (15) & 14.2 (77) \\
Irregular Range II ($-2.35$ to 7.85) & \textbf{5.2} (26) & \textbf{6.7} (8) & 3.8 (17) & 8.0 (14) & 2.0 (10) & 15.3 (83) \\
\bottomrule
\end{tabular*}
\caption{\textbf{Disagreement rates across scoring topologies.} GPT-5-mini generates the benchmark responses and evaluates the same responses under all scoring topologies. Values denote the percentage of instances whose normalized scores differ from the Likert-5 reference by more than $\delta \in \{10,20\}$, with counts shown in parentheses. Higher rates indicate greater sensitivity to numerical scoring topology.}
\label{tab:delta_statistics_pct}
\end{table*}

\begin{table}[!t]
\centering
\scriptsize
\setlength{\tabcolsep}{4pt}
\renewcommand{\arraystretch}{1}
\begin{tabular*}{\columnwidth}{
@{\extracolsep{\fill}}
lcc
@{}
}
\toprule
\textbf{Benchmark}
& Avg$_{>10}$
& Avg$_{>20}$ \\
\midrule
\multicolumn{3}{c}{\textbf{Judgment-Intensive Tasks}} \\
\midrule
Arena-Hard   & 9.9  & 2.9 \\
IFEval       & \textbf{26.0} & \textbf{13.9} \\
WildBench    & 23.1 & 8.6 \\
WritingBench & 20.7 & 6.2 \\
\cmidrule(lr){1-3}
\textit{Group Average}
& \textbf{19.9}
& \textbf{7.9} \\
\midrule
\multicolumn{3}{c}{\textbf{Answer-Verifiable Tasks}} \\
\midrule
MATH-500  & 7.3  & 2.8 \\
AIME      & 14.6 & 2.9 \\
GPQA Main & 13.4 & 2.8 \\
LCB       & 11.8 & 5.7 \\
\cmidrule(lr){1-3}
\textit{Group Average}
& 11.8
& 3.6 \\
\bottomrule
\end{tabular*}

\caption{\textbf{Average sensitivity to numerical scoring topology by task type using GPT-5-mini.}
Each value shows the disagreement rate relative to Likert-5, averaged over the twelve non-reference scoring topologies. Group averages are macro-averaged across the four benchmarks.}
\label{tab:task-type-sensitivity}
\end{table}

\begin{table*}[t]
\centering
\small
\begin{tabular*}{\textwidth}{@{\extracolsep{\fill}}llcccccccc@{}}
\toprule
Factor & Scoring Scale & MATH & AIME & GPQA & Arena & LCB & IFEval & Avg$_{>10}$ & Avg$_{>20}$ \\
\midrule
\multirow{5}{*}{Width sweep}
 & Baseline (0.74 to 1.48) & 8.0 & 14.2 & 17.2 & 13.8 & 12.0 & 30.9 & 16.0 & 6.4 \\
 & Width 0.74 to 2.48 & 9.0 & 7.5 & 17.9 & 8.4 & 10.9 & 27.7 & 13.6 & 6.7 \\
 & Width 0.74 to 4.74 & 8.0 & 6.7 & 10.3 & 4.4 & 9.1 & 20.3 & 9.8 & 4.4 \\
 & Width 0.74 to 8.74 & 7.4 & 7.5 & 10.3 & 4.6 & 7.4 & 23.8 & 10.2 & 5.1 \\
 & Width 0.74 to 15.74 & 6.8 & 4.2 & 8.7 & 4.2 & 8.6 & 25.0 & 9.6 & 4.6 \\
\midrule
\multirow{3}{*}{Round Max}
 & 1.0 to 2.0 & 9.2 & 9.2 & 17.0 & 7.6 & 14.9 & 27.2 & 14.2 & 8.8 \\
 & 3.0 to 4.0 & 8.4 & 10.0 & 19.2 & 14.4 & 10.3 & 32.9 & 15.9 & 9.7 \\
 & 6.0 to 7.0 & 9.6 & 10.8 & 17.0 & 11.6 & 14.3 & 32.7 & 16.0 & 11.3 \\
\midrule
\multirow{3}{*}{Non-round Max}
 & 1.3 to 2.04 & 8.2 & 8.3 & 15.4 & 4.8 & 9.7 & 25.9 & 12.1 & 6.2 \\
 & 3.7 to 4.7 & 36.0 & 32.5 & 65.0 & 73.4 & 75.4 & 72.8 & \textbf{59.2} & \textbf{22.7} \\
 & 6.2 to 7.2 & 63.8 & 63.3 & 81.5 & 96.6 & 91.4 & 92.4 & \textbf{81.5} & \textbf{57.7} \\
\midrule
\multirow{2}{*}{Narrow Non-round}
 & 2.11 to 2.85 & 7.6 & 11.7 & 12.7 & 10.4 & 15.4 & 29.2 & 14.5 & 7.4 \\
 & 3.33 to 4.07 & 7.0 & 10.0 & 14.5 & 6.2 & 9.1 & 27.4 & 12.4 & 6.1 \\
\bottomrule
\end{tabular*}
\caption{\textbf{Ablation analysis of boundary alignment and score width using GPT-5-mini.}
Each value denotes the percentage of instances whose normalized score differs from the Linear-100 reference by more than 10 ($\mathrm{Avg}_{>10}$) or 20 ($\mathrm{Avg}_{>20}$). MATH, GPQA, and Arena refer to MATH-500, GPQA Main, and Arena-Hard, respectively. LCB denotes LiveCodeBench. The 0.74--1.48 baseline corresponds to the shifted range used in the main experiment. Bold values highlight large increases in disagreement under non-round maximum boundaries.}
\label{tab:ablation_full}
\end{table*}

\begin{table}[t]
\centering
\small
\setlength{\tabcolsep}{7pt}
\renewcommand{\arraystretch}{1}
\begin{tabular}{lccc}
\toprule
\textbf{Scoring Range}
& \textbf{MFS}
& \textbf{Count}
& \textbf{Prop.} \\
\midrule
Likert-5 (0--5)
& 5.0
& 1,801
& 78.9\% \\

Non-round (3.7--4.7)
& 4.6
& 752
& 32.9\% \\

Non-round (6.2--7.2)
& 7.0
& 1,335
& 58.5\% \\
\bottomrule
\end{tabular}

\caption{\textbf{Concentration of raw judge scores under different scoring ranges.}
MFS denotes the most frequently assigned score, and Prop. denotes its proportion across 2,284 evaluation instances.}
\label{tab:raw-score-concentration}
\end{table}

\begin{table*}[t]
\centering
\small
\setlength{\tabcolsep}{4pt}
\renewcommand{\arraystretch}{0.95}
\begin{tabular*}{\textwidth}{@{\extracolsep{\fill}}lcccccc@{}}
\toprule
Scoring Scale & MATH-500 & AIME & GPQA Main & LCB & Arena-Hard & IFEval \\
\midrule
\multicolumn{7}{c}{Disagreement Rate ($|\Delta| > 10$)} \\
\midrule
Unit-interval (0 to 1) & \kp{11.6}{2.6} & \kn{20.0}{0.8} & \kp{21.0}{0.2} & \kp{16.6}{6.9} & \kp{6.8}{2.0} & \kp{32.5}{4.4} \\
Likert-10 (0 to 10) & \kp{6.2}{1.4} & \kn{9.2}{0.8} & \kp{12.7}{5.8} & \kp{10.9}{2.9} & \kn{5.6}{0.6} & \kp{27.9}{7.9} \\
Linear-50 (0 to 50) & \kp{5.6}{0.8} & \kp{8.3}{1.7} & \kp{12.1}{4.5} & \kp{13.1}{4.6} & \kp{9.4}{3.0} & \kp{26.8}{5.4} \\
Linear-100 (0 to 100) & \kp{7.0}{2.0} & \kn{8.3}{0.8} & \kp{12.7}{5.1} & \kp{14.3}{8.6} & \kp{8.2}{1.6} & \kp{28.1}{3.9} \\
Linear-1000 (0 to 1000) & \kp{6.4}{0.6} & \kp{9.2}{0.8} & \kp{11.6}{4.5} & \kp{14.9}{4.6} & \kp{9.0}{2.8} & \kp{28.3}{7.6} \\
Positive Offset (3 to 15) & \kp{10.4}{2.0} & \kz{17.5} & \kp{19.2}{2.2} & \kz{12.6} & \kp{11.2}{1.8} & \kp{33.6}{5.9} \\
Narrow Range (0.3 to 0.7) & \kp{14.6}{5.6} & \textbf{\kp{19.2}{10.0}} & \kp{26.8}{13.2} & \kp{23.4}{4.0} & \kp{44.0}{20.6} & \kp{47.3}{15.2} \\
Symmetric Range ($-5$ to 5) & \kp{7.6}{2.4} & \kn{15.8}{2.5} & \kp{14.7}{3.3} & \kp{12.0}{2.3} & \kp{7.0}{2.0} & \kp{29.0}{7.6} \\
Asymmetric Range ($-5$ to 2) & \kp{8.8}{0.2} & \kn{15.0}{1.7} & \kp{16.3}{4.2} & \kp{16.6}{3.4} & \kp{8.4}{1.2} & \kp{32.3}{6.7} \\
Shifted Range (0.74 to 1.48) & \textbf{\kp{22.4}{13.6}} & \kp{27.5}{4.2} & \textbf{\kp{40.4}{17.9}} & \textbf{\kp{47.4}{28.0}} & \textbf{\kp{53.8}{25.0}} & \textbf{\kp{53.6}{17.2}} \\
Irregular Range I ($-3.7$ to 8.9) & \kp{13.4}{4.6} & \kp{19.2}{0.8} & \kp{20.8}{4.9} & \kp{13.1}{1.1} & \kp{12.2}{4.6} & \kp{33.1}{7.6} \\
Irregular Range II ($-2.35$ to 7.85) & \kp{13.2}{3.6} & \kp{20.8}{4.2} & \kp{18.5}{0.2} & \kp{16.0}{2.9} & \kp{11.6}{4.0} & \kp{34.2}{5.4} \\
\midrule
\multicolumn{7}{c}{Disagreement Rate ($|\Delta| > 20$)} \\
\midrule
Unit-interval (0 to 1) & \kp{4.4}{1.8} & \kz{3.3} & \kp{5.4}{2.5} & \kp{8.0}{2.3} & \kp{2.6}{0.8} & \kp{19.6}{4.4} \\
Likert-10 (0 to 10) & \kp{3.6}{2.2} & \kn{0.8}{0.8} & \kp{4.0}{2.5} & \kp{6.3}{1.7} & \kp{2.0}{1.0} & \kp{15.5}{4.3} \\
Linear-50 (0 to 50) & \kp{2.6}{1.4} & \kp{3.3}{2.5} & \kp{4.5}{2.5} & \kp{8.0}{3.4} & \kp{3.0}{1.2} & \kp{15.2}{2.8} \\
Linear-100 (0 to 100) & \kp{3.4}{1.2} & \kn{2.5}{0.8} & \kp{4.2}{3.1} & \kp{8.6}{5.7} & \kp{2.6}{1.6} & \kp{15.0}{2.4} \\
Linear-1000 (0 to 1000) & \kz{2.8} & \kp{4.2}{2.5} & \kp{4.2}{2.5} & \kp{9.7}{3.4} & \kp{3.0}{0.2} & \kp{16.5}{4.1} \\
Positive Offset (3 to 15) & \kp{4.6}{2.2} & \kp{3.3}{0.8} & \kp{5.4}{1.6} & \kp{6.9}{1.7} & \kp{4.4}{1.0} & \kp{17.7}{1.8} \\
Narrow Range (0.3 to 0.7) & \kp{5.6}{3.8} & \kp{3.3}{1.7} & \kp{6.5}{4.5} & \kn{7.4}{1.1} & \kn{6.0}{1.8} & \kp{22.7}{4.6} \\
Symmetric Range ($-5$ to 5) & \kp{3.6}{0.8} & \kp{5.0}{1.7} & \kp{4.5}{2.7} & \kp{8.0}{1.7} & \kp{2.2}{0.6} & \kp{18.3}{5.9} \\
Asymmetric Range ($-5$ to 2) & \kp{4.4}{0.2} & \kp{4.2}{1.7} & \kp{7.1}{1.8} & \kp{6.9}{2.3} & \kn{2.4}{0.2} & \kp{19.4}{5.9} \\
Shifted Range (0.74 to 1.48) & \textbf{\kp{12.8}{10.4}} & \textbf{\kp{9.2}{4.2}} & \textbf{\kp{12.5}{9.6}} & \textbf{\kp{21.7}{14.9}} & \textbf{\kp{13.4}{7.0}} & \textbf{\kp{29.4}{15.7}} \\
Irregular Range I ($-3.7$ to 8.9) & \kp{7.6}{3.2} & \kp{5.8}{3.3} & \kp{9.2}{4.2} & \kp{8.0}{2.9} & \kp{3.2}{0.2} & \kp{18.5}{4.3} \\
Irregular Range II ($-2.35$ to 7.85) & \kp{7.6}{2.4} & \kp{8.3}{1.7} & \kp{6.9}{3.1} & \kp{10.9}{2.9} & \kp{2.2}{0.2} & \kp{19.2}{3.9} \\
\bottomrule
\end{tabular*}

\caption{\textbf{Cross-lingual impact on evaluator calibration.} Disagreement rates under Korean-language judging relative to the condition-specific Likert-5 reference. Values in parentheses denote percentage-point differences compared with English-language judging (red: increased instability, blue: decreased instability). The results indicate that Korean-language evaluation can increase sensitivity to numerical scoring topology.}

\label{tab:korean_judge_delta_statistics}
\end{table*}

\begin{table}[t]
\centering
\small
\setlength{\tabcolsep}{5pt}
\renewcommand{\arraystretch}{1}

\begin{tabular}{lccccc}
\toprule
\textbf{Condition}
& \textbf{MATH}
& \textbf{LCB}
& \textbf{Arena}
& \textbf{IFEval}
& \textbf{Max} \\
\midrule

EN$\rightarrow$EN
& 7.5
& \textbf{12.0}
& 10.2
& \textbf{26.3}
& 33.9 \\

KO$\rightarrow$EN
& \textbf{1.8}
& 13.0
& \textbf{9.3}
& 29.9
& 31.9 \\

EN$\rightarrow$KO
& 10.6
& 17.6
& 15.6
& 33.9
& 21.9 \\

KO$\rightarrow$KO
& 4.1
& 14.9
& 14.4
& 39.0
& 23.0 \\

EN$\rightarrow$ZH
& 4.2
& 16.8
& 17.6
& 33.5
& 14.3 \\

EN$\rightarrow$JA
& 2.1
& 15.3
& 13.8
& 34.2
& \textbf{38.7} \\

\bottomrule
\end{tabular}

\caption{\textbf{Disagreement rates and maximum-score usage across language conditions using GPT-5-mini.} The benchmark columns show $|\Delta|>10$ disagreement rates averaged across twelve non-reference topologies. Max shows maximum-score usage for the shifted range 0.74--1.48 across four benchmarks. EN$\rightarrow$KO, EN$\rightarrow$ZH and EN$\rightarrow$JA denote English responses evaluated using Korean, Chinese, and Japanese judge prompts, respectively.}
\label{tab:language-average-disagreement}
\end{table}

% \begin{table}[t]
% \centering
% \small
% \setlength{\tabcolsep}{6pt}
% \renewcommand{\arraystretch}{1.08}
% \begin{tabular}{lccccc}
% \toprule
% \textbf{Condition}
% & \textbf{MATH}
% & \textbf{LCB}
% & \textbf{Arena}
% & \textbf{IFEval}
% & \textbf{Max} \\
% \midrule
% EN$\rightarrow$EN
% & 7.5
% & \textbf{12.0}
% & 10.2
% & \textbf{26.3}
% & \textbf{33.9} \\

% KO$\rightarrow$EN
% & \textbf{1.8}
% & 13.0
% & \textbf{9.3}
% & 29.9
% & 31.9 \\

% EN$\rightarrow$KO
% & 10.6
% & 17.6
% & 15.6
% & 33.9
% & 21.9 \\

% KO$\rightarrow$KO
% & 4.1
% & 14.9
% & 14.4
% & 39.0
% & 23.0 \\
% \bottomrule
% \end{tabular}

% \caption{\textbf{Disagreement rates and maximum-score usage across language conditions using GPT-5-mini.}
% The benchmark columns show $|\Delta|>10$ disagreement rates averaged across twelve non-reference topologies. Max shows maximum-score usage for the shifted range 0.74--1.48 across four benchmarks. Lower disagreement and higher Max indicate greater stability.}
% \label{tab:language-average-disagreement}
% \end{table}

\section{Experimental Setup}

Due to space limitations, detailed descriptions of the scoring topologies, evaluation metrics, and inference settings are provided in Appendix~\ref{appendix:experimental-setup}.

We use an experimental framework in which only the numerical scoring topology varies, while the evaluation task, prompt, criteria, model output, and judge model remain fixed. This setup isolates the effect of score design on evaluator behavior.

We conduct the primary thirteen-topology analysis on six benchmarks and extend the task-type analysis to two additional judgment-intensive benchmarks, using seventeen LLM judges in the generalization experiments.

\subsection{Evaluation Benchmarks}

We evaluate LLM judges on eight benchmarks covering mathematical reasoning, scientific reasoning, code generation, conversational quality, instruction following, and writing evaluation.

% We evaluate LLM judges on six representative benchmarks spanning both objective reasoning and subjective quality assessment.

\textbf{MATH-500}~\cite{lightman2024let} and \textbf{AIME}~\footnote{\url{https://huggingface.co/datasets/allenai/aime-2022-2025}} evaluate mathematical reasoning with well-defined reference answers. We use 500 examples from MATH-500 and 120 problems from AIME 2022--2025. \textbf{GPQA Main}~\cite{rein2023gpqa} contains 448 expert-level questions in biology, physics, and chemistry. \textbf{LiveCodeBench (LCB)}~\cite{jain2024livecodebench} includes 175 code-generation examples spanning easy, medium, and hard difficulty levels.

\textbf{Arena-Hard}~\cite{li2024crowdsourced} contains 500 challenging prompts, while \textbf{WildBench}~\cite{lin2024wildbench} includes 256 real-world user queries. Both benchmarks evaluate response quality in settings where judges assess dimensions such as helpfulness, relevance, and overall quality. \textbf{IFEval}~\cite{zhou2023instruction} contains 541 examples with explicit instruction-following constraints. \textbf{WritingBench}~\cite{wu2026writingbench} evaluates generated writing across diverse tasks and criteria, and we use 555 examples from its English subset.

% \textbf{MATH} and \textbf{AIME} evaluate mathematical reasoning using deterministic ground-truth answers, where evaluation depends primarily on objective correctness. \textbf{GPQA} measures expert-level scientific reasoning and factual knowledge, while \textbf{LiveCodeBench (LCB)} evaluates code generation through executable correctness.

% In contrast, \textbf{Chatbot Arena} measures preference-based conversational quality through human-aligned judgments, and \textbf{IFEval} evaluates a model's ability to satisfy complex instruction-following constraints. Compared with reasoning benchmarks, these tasks require substantially greater evaluator discretion when assigning numerical scores.

These benchmarks represent evaluation settings that differ in answer verifiability and in the degree of evaluator discretion required when assigning numerical scores. This diversity allows us to examine whether sensitivity to numerical scoring topology varies across different forms of evaluation.

% Together, these benchmarks span a broad spectrum of evaluation settings, allowing us to examine whether sensitivity to numerical scoring topology varies according to the degree of evaluation subjectivity.

%%%%%%%%%%%%%%%%%%%%%%%%%%%%%%%%%%%%%%%%%%%%%%%%%%%%

\subsection{Judge Models}
We use GPT-5-mini as the primary model for both inference and judging. To examine whether sensitivity to numerical scoring topology generalizes across model families and scales, we evaluate seventeen LLM judge models: GPT-5.4~\footnote{\url{https://openai.com/index/introducing-gpt-5-4/}}, GPT-5-mini~\footnote{\url{https://openai.com/index/gpt-5-system-card/}}, GPT-4o-mini~\cite{hurst2024gpt}, GPT-OSS-120B, GPT-OSS-20B~\cite{openai2025gptoss120bgptoss20bmodel}, MiniMax-M3~\footnote{\url{https://www.minimax.io/blog/minimax-m3}}, DeepSeek-V4-Pro~\cite{deepseekai2026deepseekv4}, Gemini-2.5-Flash-Lite~\cite{comanici2025gemini}, Qwen3.7-Plus, Qwen3-32B, Qwen3-14B, Qwen3-8B~\cite{qwen3technicalreport}, Llama-3.1-70B, Llama-3.1-8B~\cite{grattafiori2024llama}, Gemma-3-27B, Gemma-3-12B, and Gemma-3-4B~\cite{gemmateam2025gemma3technicalreport}. The complete generation and judging prompts, including multilingual variants and prompt-based mitigation strategies, are provided in Appendix~\ref{appendix:prompts}.

\subsection{Experimental Design}
Our experimental design follows the five research questions introduced in Section~\ref{sec:introduction}.

For \textbf{RQ1}, we use GPT-5-mini as both the response generator and judge model and evaluate identical responses across thirteen scoring topologies. We compare normalized scores and disagreement rates across benchmarks to examine how numerical scoring topology affects evaluation outcomes and task-level sensitivity.

For \textbf{RQ2}, we conduct ablations of score width and boundary configuration. We vary range width, boundary placement, numerical offset, and boundary alignment while holding the benchmark responses, evaluation criteria, and judge model fixed.

For \textbf{RQ3}, we compare six response--evaluation language configurations: EN$\rightarrow$EN, KO$\rightarrow$EN, EN$\rightarrow$KO, KO$\rightarrow$KO, EN$\rightarrow$ZH, and EN$\rightarrow$JA. This setup examines the effects of response language, evaluation language, and their interaction on sensitivity to numerical scoring topology.

For \textbf{RQ4}, we evaluate response generators and seventeen judge models across multiple model families and scales. We measure judge discrimination using AUROC and topology robustness using disagreement rates to assess whether the observed effects generalize across generators and judges.

For \textbf{RQ5}, we compare two prompt-based mitigation strategies, Allowed List and Explicit Max, against a scale-only baseline. We measure disagreement rates relative to the Likert-5 reference across all conditions and maximum-score usage for the two mitigation strategies to assess their effectiveness in reducing sensitivity to numerical scoring topology.

\section{Results}
% We organize our analysis according to the three research questions introduced in Section~\ref{sec:introduction}. Across all experiments, we find that numerical scoring topology consistently influences LLM-as-a-Judge evaluation. However, its impact is not uniform: the magnitude of the effect varies depending on the evaluation task, the numerical boundary configuration, and the evaluation language. These findings indicate that seemingly minor design choices in numerical scoring schemes can substantially affect evaluation outcomes.

%%%%%%%%%%%%%%%%%%%%%%%%%%%%%%%%%%%%%%%%%%%%%%%%%%%%

\subsection{RQ1: How does numerical scoring topology influence LLM-as-a-Judge evaluation across different scoring schemes and evaluation tasks?}
% \subsection{RQ1: How Does Scoring Topology Affect Different Evaluation Tasks?}

We examine the overall effect of numerical scoring topology, identify the numerical properties responsible for evaluation instability, and compare sensitivity across task types.

% We first examine whether changing the numerical scoring topology alters evaluation outcomes under otherwise identical conditions. Table~\ref{tab:scoring-scale-results} presents the normalized evaluation scores obtained under the thirteen scoring topologies, while Table~\ref{tab:delta_statistics_pct} shows the corresponding disagreement rates relative to the Likert-5 reference. We then conduct controlled ablation studies to identify which numerical properties, including score width, boundary placement, and boundary alignment, contribute most strongly to evaluation instability. Finally, we compare tasks with different degrees of verifiable correctness and evaluator discretion to analyze how task characteristics affect sensitivity to numerical scoring topology.

% We first investigate whether the influence of scoring topology depends on the evaluation task. Table~\ref{tab:scoring-scale-results} summarizes evaluation performance under the thirteen scoring topologies, while Table~\ref{tab:delta_statistics_pct} reports disagreement rates after score normalization. Figure~\ref{fig:judge-behavior} further analyzes how different judge models utilize the available scoring space under each topology.

\paragraph{Finding 1: Numerical scoring topology changes evaluation outcomes even when all semantic evaluation conditions are held constant.}

% \paragraph{Finding 1: Subjective evaluation tasks are substantially more sensitive to scoring topology than objective reasoning tasks.}

Table~\ref{tab:scoring-scale-results} shows that normalized benchmark scores vary when only the numerical scoring topology is changed. The largest decreases occur under the Shifted Range, including $-7.5$ points on Arena-Hard and $-5.6$ points on IFEval, whereas several asymmetric and irregular ranges increase scores on other benchmarks. This indicates that topology changes affect both the magnitude and direction of evaluation outcomes.

% Objective reasoning benchmarks such as MATH, AIME, GPQA, and LiveCodeBench remain remarkably stable across most scoring topologies. Their normalized disagreement rates remain consistently low even under shifted or irregular numerical ranges, indicating that changing the numerical scale rarely alters evaluator decisions when correctness is largely deterministic.

Table~\ref{tab:delta_statistics_pct} further confirms substantial instance-level variation. Across all non-reference topologies, 14.2\% of matched evaluation instances differ from Likert-5 by more than 10 normalized points, and 5.7\% differ by more than 20 points. The Shifted Range produces the highest overall disagreement at $|\Delta|>10$, reaching 24.0\%, while the Narrow Range produces the highest disagreement for $|\Delta|>20$ at 7.5\%. These results demonstrate that numerical score design itself systematically alters LLM judge evaluations.

\paragraph{Finding 2: Judgment-intensive tasks are more sensitive to numerical scoring topology than answer-verifiable tasks.}
The task-level variation observed in Tables~\ref{tab:scoring-scale-results} and~\ref{tab:delta_statistics_pct} motivates an additional analysis of evaluation type. We therefore extend the benchmark suite with WildBench and WritingBench and group the tasks into answer-verifiable benchmarks (MATH-500, AIME, GPQA Main, and LCB) and judgment-intensive benchmarks (Arena-Hard, IFEval, WildBench, and WritingBench). Detailed results for WildBench and WritingBench are provided in
Appendix~\ref{appendix:open-ended-scores} and Appendix~\ref{appendix:open-ended-disagreement}.

As shown in Table~\ref{tab:task-type-sensitivity}, the judgment-intensive group exhibits higher disagreement rates than the answer-verifiable group: 19.9\% versus 11.8\% for $|\Delta|>10$ and 7.9\% versus 3.6\% for $|\Delta|>20$. IFEval, WildBench, and WritingBench show particularly high sensitivity, whereas Arena-Hard remains relatively stable. These results indicate that tasks requiring qualitative assessment or the aggregation of multiple criteria are generally more sensitive to numerical score representation than tasks with explicit answers or execution-based correctness.

% \paragraph{Finding 3: Sensitivity to numerical scoring topology varies across evaluation tasks.}

% Figure~\ref{fig:judge-behavior}(c) compares the variability among judge models under different scoring topologies. Fine-grained linear scales (0--50, 0--100, and 0--1000) produce the largest between-model variation, whereas binary, narrow, and asymmetric scales lead to substantially smaller differences across judges.

% This result indicates that numerical scoring topology not only affects individual judge calibration but also influences the extent to which different LLM judges agree with one another. Consequently, selecting an appropriate scoring topology is important for obtaining reproducible evaluations across model families.

%%%%%%%%%%%%%%%%%%%%%%%%%%%%%%%%%%%%%%%%%%%%%%%%%%%%

\subsection{RQ2: Which properties of numerical scoring topology drive evaluation instability?}
% \subsection{RQ2: How does the interaction between response language and evaluation language affect sensitivity to numerical scoring topology?}
% \subsection{RQ2: How Do Numerical Boundaries Affect Evaluator Calibration?}

We analyze how score width and boundary configuration affect evaluation stability and examine the raw-score behavior underlying highly unstable configurations.

\paragraph{Finding 1: Boundary configuration affects evaluation stability more strongly than score width alone.}

Table~\ref{tab:ablation_full} compares the effects of score width and boundary configuration. Increasing the width of the shifted range generally reduces disagreement, with the average $|\Delta|>10$ rate decreasing from 16.0\% under 0.74--1.48 to 9.6\% under 0.74--15.74. A similar overall tendency is observed at $|\Delta|>20$, although the reduction is not strictly monotonic. These results suggest that wider scoring ranges generally reduce sensitivity to numerical scoring topology.

% The width-controlled experiments show that broader numerical ranges moderately reduce disagreement. As the upper boundary increases from 1.48 to 15.74 while maintaining the same lower boundary, the average disagreement rate decreases from 16.0\% to 9.6\%.

Boundary configuration produces substantially larger effects. Ranges with round maximum values show average $|\Delta|>10$ disagreement rates between 14.2\% and 16.0\%, whereas 3.7--4.7 and 6.2--7.2 increase disagreement to 59.2\% and 81.5\%, respectively. However, the 1.3--2.04 range remains comparatively stable because judges frequently assign a score of 2.0, which lies close to the upper boundary of 2.04 and therefore maps to a high normalized score. Thus, instability is associated with how frequently used numerical anchors align with the range boundaries, rather than with non-round values in general. 

% Although this trend suggests that wider score spaces allow judges to distribute scores more consistently, the improvement remains relatively modest. Therefore, score width alone cannot explain the substantial behavioral differences observed across scoring topologies.

\paragraph{Finding 2: Unstable boundary configurations induce concentration around score anchors that do not align with the specified maximum.}

To investigate the mechanism underlying the high disagreement, we analyze the raw scores assigned under the two most unstable configurations. Table~\ref{tab:raw-score-concentration} shows the most frequently assigned score (MFS) and its proportion. For Likert-5, the judge assigns the maximum score of 5.0 in 78.9\% of evaluations. In contrast, scores in the non-round ranges concentrate on values below their specified maxima.

% A considerably larger effect emerges when only the numerical boundary values are changed. Scoring schemes with round-number upper bounds (e.g., 2.0, 4.0, and 7.0) consistently exhibit disagreement rates comparable to conventional evaluation scales. In contrast, replacing these values with nearby non-round boundaries produces dramatic increases in disagreement despite minimal numerical changes.

For 3.7--4.7, the most frequent score is 4.6, accounting for 32.9\% of the evaluations. For 6.2--7.2, the judge assigns 7.0 in 58.5\% of the evaluations rather than using the specified maximum of 7.2. After normalization, these values correspond to 90 and 80 points, respectively, whereas the most frequent Likert-5 score corresponds to 100 points. Consequently, judgments concentrated near the upper end of the raw scale can be mapped to substantially different normalized scores depending on boundary placement.

These results suggest that the judge does not consistently adapt its preferred score anchors to the exact boundaries of the scoring range. The mismatch between raw-score concentration and the specified boundaries provides a behavioral explanation for the large disagreement observed under specific numerical configurations. Detailed boundary-ablation results are provided in Appendix~\ref{appendix:boundary-ablation-results} and Appendix~\ref{appendix:raw-score-concentration}.

% For example, changing the upper boundary from 7.0 to 7.2 increases the average disagreement rate from approximately 16\% to 81.5\%, while the 3.7--4.7 configuration reaches 59.2\%. These differences are substantially larger than those observed in the width-controlled experiments, indicating that boundary placement plays a dominant role in evaluator calibration.

% \paragraph{Finding 3: Calibration instability cannot be attributed solely to narrow score intervals.}

% The narrow non-round configurations provide additional evidence that instability is not simply caused by reducing the available score range. Despite having similarly narrow widths, these configurations remain considerably more stable than several non-round maximum settings.

% Taken together, these controlled experiments consistently indicate that the numerical values defining score boundaries exert a stronger influence on evaluator behavior than the overall width of the scoring scale. This suggests that LLM judges are sensitive not only to the amount of available scoring resolution but also to the specific numerical anchors presented during evaluation.

%%%%%%%%%%%%%%%%%%%%%%%%%%%%%%%%%%%%%%%%%%%%%%%%%%%%%

\subsection{RQ3: How does the interaction between response language and evaluation language affect sensitivity to numerical scoring topology?}

We analyze whether sensitivity to numerical scoring topology changes across response and evaluation languages. By comparing six language configurations, we examine whether the effect of cross-lingual evaluation varies across tasks.

\paragraph{Finding 1: Korean evaluation increases sensitivity to numerical scoring topology.}

Table~\ref{tab:korean_judge_delta_statistics} compares disagreement rates under EN$\rightarrow$KO condition with those under the EN$\rightarrow$EN baseline. Across most benchmarks and scoring topologies, Korean evaluation produces higher disagreement relative to the condition-specific Likert-5 reference. The increase is particularly pronounced for the Shifted Range, where the $|\Delta|>10$ disagreement rate rises by 13.6 percentage points on MATH-500, 28.0 points on LCB, 25.0 points on Arena-Hard, and 17.2 points on IFEval.

The effect is also visible under the stricter $|\Delta|>20$ threshold. For the Shifted Range, disagreement increases by 10.4 percentage points on MATH-500, 14.9 points on LCB, 7.0 points on Arena-Hard, and 15.7 points on IFEval. Although some configurations show reduced or unchanged disagreement, the overall pattern indicates that numerical scoring topology becomes more sensitive when English responses are evaluated using Korean instructions.

\paragraph{Finding 2: The effect of language configuration varies across tasks and evaluation languages.}

Table~\ref{tab:language-average-disagreement} shows that disagreement rates vary substantially across language conditions. English evaluation is generally the most stable, with KO$\rightarrow$EN achieving the lowest disagreement on MATH-500 and Arena-Hard and EN$\rightarrow$EN performing best on LCB and IFEval. However, the effect of non-English evaluation is not uniform. Japanese evaluation remains relatively stable on MATH-500, reaching 2.1\% disagreement, but produces higher disagreement on LCB, Arena-Hard, and IFEval. Chinese evaluation also shows greater sensitivity on the remaining benchmarks. Korean evaluation produces consistently elevated disagreement, with IFEval reaching 39.0\% under KO$\rightarrow$KO.

% Table~\ref{tab:language-average-disagreement} shows that the average disagreement rate varies substantially across language conditions. KO$\rightarrow$EN is the most stable condition, with an overall average of 13.5\%, while EN$\rightarrow$KO and KO$\rightarrow$KO increase to 19.4\% and 18.1\%, respectively. The variation is also task dependent: MATH-500 ranges from 1.8\% to 10.6\%, whereas IFEval exhibits the highest disagreement across all conditions, reaching 39.0\% under KO$\rightarrow$KO.

These results indicate that Korean evaluation is consistently sensitive to scoring topology. Evaluation language also affects maximum-score usage under the shifted range: it decreases from 31.9--33.9\% under English evaluation to 21.9--23.0\% under Korean evaluation and 14.3\% under Chinese evaluation, whereas Japanese evaluation reaches 38.7\%. This suggests that adaptation to the upper boundary depends on the evaluation language. Detailed results are provided in Appendix~\ref{appendix:language-condition-results}.

% These results indicate that Korean evaluation is more sensitive to scoring topology, while Japanese and Chinese evaluation exhibit more task-dependent patterns. This is also reflected in maximum-score usage under the shifted range. Maximum-score usage decreases from 31.9--33.9\% under English evaluation to 21.9--23.0\% under Korean evaluation and further to 14.3\% under Chinese evaluation, whereas Japanese evaluation reaches the highest value at 38.7\%. These differences suggest that adaptation to the specified upper boundary depends on the evaluation language, with Korean and Chinese prompts showing weaker alignment with the shifted maximum and Japanese prompts preserving stronger upper-bound usage. Detailed results across response--evaluation language conditions are provided in Appendix~\ref{appendix:language-condition-results}.

\begin{table}[t]
\centering
\small
\setlength{\tabcolsep}{5.5pt}
\renewcommand{\arraystretch}{1}
\begin{tabular}{lccc}
\toprule
\textbf{Judge Model}
& \textbf{AUROC}
& \textbf{$|\Delta|>10$}
& \textbf{$|\Delta|>20$} \\
\midrule
GPT-5.4
& \textbf{0.934} & \textbf{12.7} & \textbf{4.9} \\

MiniMax-M3
& 0.931 & 17.5 & 10.6 \\

GPT-OSS-120B
& 0.930 & 16.7 & \textbf{2.5} \\

DeepSeek-V4-Pro
& 0.919 & 13.7 & 12.1 \\

GPT-5-mini
& 0.913 & 22.7 & 3.3 \\

Qwen3.7-Plus
& 0.907 & \textbf{5.3} & 5.3 \\
\midrule
Gemini-2.5-Flash-Lite
& 0.900 & 38.7 & 19.0 \\

Gemma-3-4B
& 0.875 & 88.0 & 86.1 \\

Qwen3-8B
& 0.841 & 15.7 & 14.1 \\

Llama-3.1-8B
& 0.696 & 99.1 & 92.6 \\
\bottomrule
\end{tabular}

\caption{\textbf{Judge discriminative validity and sensitivity to numerical scoring topology.} AUROC is computed on pooled GPQA Main and AIME, while $|\Delta|>10$ and $|\Delta|>20$ report cross-topology disagreement rates.}
\label{tab:frontier-judge-topology}
\end{table}

\subsection{RQ4: How consistently do the effects of numerical scoring topology generalize across different response generators and LLM judge models?}

We use GPQA Main and AIME because both provide objectively verifiable answers, enabling generated responses to be labeled as correct or incorrect independently of the judge. This allows us to measure judge discrimination using AUROC. We additionally measure topology sensitivity using $|\Delta|>10$ and $|\Delta|>20$, where $\Delta$ is the normalized-score difference between Likert-5 and the Shifted Range (0.74--1.48).

\paragraph{Finding 1: Higher-capability judges generally combine discriminative validity with greater robustness to numerical scoring topology.}

Table~\ref{tab:frontier-judge-topology} compares six high-capability judges with four lower-capability judges using responses generated by GPT-5-mini. Higher-capability judges achieve AUROC values between 0.907--0.934 on GPQA Main and AIME. Among the lower-capability judges, Gemini-2.5-Flash-Lite obtains an AUROC of 0.900, followed by Gemma-3-4B at 0.875, Qwen3-8B at 0.841, and Llama-3.1-8B at 0.696.

Higher-capability judges also tend to be less sensitive to numerical scoring topology. Qwen3.7-Plus obtains the lowest $|\Delta|>10$ disagreement rate at 5.3\%, while GPT-OSS-120B obtains the lowest $|\Delta|>20$ rate at 2.5\%. By contrast, Gemma-3-4B and Llama-3.1-8B exhibit disagreement rates above 86\% at both thresholds. Overall, higher judge capability is generally associated with strong discriminative validity and cross-topology robustness.

\begin{table}[t]
\centering
\small
\setlength{\tabcolsep}{5.2pt}
\renewcommand{\arraystretch}{1}
\begin{tabular}{lccc}
\toprule
\textbf{Judge Model}
& \textbf{AUROC}
& \textbf{$|\Delta|>10$}
& \textbf{$|\Delta|>20$} \\
\midrule
GPT-5.4
& \textbf{0.934} & \textbf{12.7} & \textbf{4.9} \\
GPT-5-mini
& 0.913 & 22.7 & 3.3 \\
GPT-4o-mini
& 0.898 & 25.0 & 7.6 \\
GPT-OSS-120B
& \textbf{0.930} & \textbf{16.7} & \textbf{2.5} \\
GPT-OSS-20B
& 0.916 & 51.1 & 10.2 \\

\midrule
Llama-3.1-70B
& \textbf{0.888} & \textbf{94.0} & \textbf{84.0} \\
Llama-3.1-8B
& 0.696 & 99.1 & 92.6 \\

\midrule
Gemma-3-27B
& \textbf{0.879} & \textbf{51.6} & \textbf{45.1} \\
Gemma-3-12B
& \textbf{0.879} & 72.7 & 50.5 \\
Gemma-3-4B
& 0.875 & 88.0 & 86.1 \\

\midrule
Qwen3-32B
& 0.841 & 21.0 & 17.8 \\
Qwen3-14B
& \textbf{0.848} & 19.5 & 15.3 \\
Qwen3-8B
& 0.841 & \textbf{15.7} & \textbf{14.1} \\

\bottomrule
\end{tabular}

\caption{\textbf{Within-group judge performance and topology sensitivity.} AUROC is computed on pooled GPQA Main and AIME, while $|\Delta|>10$ and $|\Delta|>20$ report cross-topology disagreement rates.}
\label{tab:grouped-judge-topology}
\end{table}

\begin{table*}[t]
\centering
\small
\setlength{\tabcolsep}{7pt}
\renewcommand{\arraystretch}{0.96}
\begin{tabular}{llccccc}
\toprule
\multirow{2}{*}{\textbf{Benchmark}}
& \multirow{2}{*}{\textbf{Scoring Range}}
& \textbf{Baseline}
& \multicolumn{2}{c}{\textbf{Allowed List}}
& \multicolumn{2}{c}{\textbf{Explicit Max}} \\
\cmidrule(lr){3-3}
\cmidrule(lr){4-5}
\cmidrule(lr){6-7}
& & $\boldsymbol{|\Delta|>10}$
  & \textbf{\%Max} & $\boldsymbol{|\Delta|>10}$
  & \textbf{\%Max} & $\boldsymbol{|\Delta|>10}$ \\
\midrule
\multirow{3}{*}{MATH-500}
& [6.2, 7.2]   & 64.8 & 72.4 & 25.4 & 95.4 & \textbf{3.2} \\
& [3.7, 4.7]   & 33.3 & 94.3 & 3.7  & 96.5 & \textbf{2.5} \\
& [0.74, 1.48] & 8.8  & 95.7 & 2.1  & 97.0 & \textbf{2.0} \\
\midrule
\multirow{3}{*}{Arena-Hard}
& [6.2, 7.2]   & 97.1 & 3.0  & 93.7 & 24.2 & \textbf{71.6} \\
& [3.7, 4.7]   & 94.5 & 14.1 & 83.9 & 35.2 & \textbf{62.8} \\
& [0.74, 1.48] & 28.8 & 71.7 & 24.6 & 63.3 & \textbf{14.1} \\
\midrule
\multirow{3}{*}{LiveCodeBench}
& [6.2, 7.2]   & 93.7 & 11.4 & 81.1 & 65.7 & \textbf{28.6} \\
& [3.7, 4.7]   & 86.9 & 42.3 & 50.9 & 64.6 & \textbf{24.6} \\
& [0.74, 1.48] & 19.4 & 78.3 & 15.4 & 78.3 & \textbf{12.6} \\
\midrule
\multirow{3}{*}{IFEval}
& [6.2, 7.2]   & 92.9 & 5.2  & 83.4 & 37.3 & \textbf{54.3} \\
& [3.7, 4.7]   & 85.5 & 16.6 & 74.9 & 43.3 & \textbf{49.6} \\
& [0.74, 1.48] & 36.4 & 60.1 & 36.0 & 60.8 & \textbf{30.6} \\
\midrule
\multirow{3}{*}{GPQA Main}
& [6.2, 7.2]   & 83.5 & 21.9 & 67.0 & 65.4 & \textbf{22.5} \\
& [3.7, 4.7]   & 72.1 & 48.9 & 41.7 & 65.6 & \textbf{21.2} \\
& [0.74, 1.48] & 22.5 & 70.2 & 23.7 & 71.9 & \textbf{18.8} \\
\midrule
\multirow{3}{*}{AIME}
& [6.2, 7.2]   & 63.3 & 52.5 & 25.8 & 70.0 & \textbf{14.2} \\
& [3.7, 4.7]   & 41.7 & 67.5 & 16.7 & 70.0 & \textbf{13.3} \\
& [0.74, 1.48] & 23.3 & 71.7 & 11.7 & 70.8 & \textbf{11.7} \\
\bottomrule
\end{tabular}

\caption{\textbf{Effectiveness of prompt-based mitigation strategies across six benchmarks and three scoring ranges using GPT-5-mini.} \%Max denotes maximum-score usage, while $|\Delta|>10$ represents the disagreement rate relative to the Likert-5 reference.}

\label{tab:mitigation}
\end{table*}

\paragraph{Finding 2: Increasing model scale generally improves judge performance and robustness to numerical scoring topology.}

% Table~\ref{tab:grouped-judge-topology} shows an overall tendency for larger models to achieve stronger discriminative validity and lower cross-topology disagreement. Within GPT-OSS, scaling from 20B to 120B increases AUROC from 0.916 to 0.930, while reducing the $|\Delta|>10$ and $|\Delta|>20$ rates from 51.1\% and 10.2\% to 16.7\% and 2.5\%, respectively. Llama-3.1 similarly improves from an AUROC of 0.696 at 8B to 0.888 at 70B, accompanied by modest reductions in both disagreement rates. Gemma-3 shows nearly unchanged AUROC across scales, but its robustness improves substantially: the $|\Delta|>20$ rate decreases from 86.1\% at 4B to 45.1\% at 27B.

Table~\ref{tab:grouped-judge-topology} shows that larger models generally achieve stronger discriminative validity and lower cross-topology disagreement. Within GPT-OSS, scaling from 20B to 120B increases AUROC from 0.916 to 0.930 while reducing the $|\Delta|>10$ and $|\Delta|>20$ rates from 51.1\% and 10.2\% to 16.7\% and 2.5\%. Llama-3.1 improves from 0.696 at 8B to 0.888 at 70B, with modest reductions in both disagreement rates. Gemma-3 shows nearly unchanged AUROC across scales, while its $|\Delta|>20$ rate decreases from 86.1\% at 4B to 45.1\% at 27B.

% Table~\ref{tab:grouped-judge-topology} shows that larger models generally achieve stronger discriminative validity and lower cross-topology disagreement. Within GPT-OSS, scaling from 20B to 120B increases AUROC from 0.916 to 0.930 while reducing the $|\Delta|>10$ and $|\Delta|>20$ rates from 51.1\% and 10.2\% to 16.7\% and 2.5\%. Llama-3.1 improves from 0.696 at 8B to 0.888 at 70B, with modest disagreement reductions. Gemma-3 shows similar AUROC across scales, while its $|\Delta|>20$ rate decreases from 86.1\% at 4B to 45.1\% at 27B.

This scaling benefit is not strictly monotonic across all model families. Within Qwen3, the 14B model achieves the highest AUROC, whereas the 8B model exhibits the lowest disagreement rates. Nevertheless, the overall results indicate that increasing parameter scale generally strengthens judge discrimination and, in several model families, reduces sensitivity to numerical scoring topology. Detailed judge-level robustness results are provided in Appendix~\ref{appendix:judge-robustness-results}, Appendix~\ref{appendix:ranking-stability}, and Appendix~\ref{appendix:response-generator-score-shifts}.

\subsection{RQ5: How effectively can prompt-based strategies mitigate evaluation instability induced by numerical scoring topology?}

We compare two prompt-based mitigation strategies, Allowed List and Explicit Max, against a baseline condition. The baseline condition asks the judge to assign a single score within the specified range. The Allowed List condition explicitly provides a predefined set of permitted score values, such as ${6.2,6.3,\ldots,7.2}$, and asks the judge to select one of them. The Explicit Max condition instructs the judge to assign the specified maximum score to a fully correct or excellent response.

% We compare three prompting conditions, including Scale-only, Allowed List, and Explicit Max. The Scale-only condition asks the judge to assign a single score within the specified range. The Allowed List condition explicitly enumerates all valid score values, such as ${6.2,6.3,\ldots,7.2}$, and asks the judge to select one of them. The Explicit Max condition instructs the judge to assign the specified maximum score to a fully correct or excellent response.

\paragraph{Finding 1: Explicit maximum-score instructions substantially reduce evaluation instability across scoring topologies.}

Table~\ref{tab:mitigation} shows that Explicit Max consistently achieves lower disagreement rates than the baseline across all evaluated benchmarks and scoring ranges. On MATH-500, the $|\Delta|>10$ disagreement rate decreases from 64.8\% to 3.2\% for $[6.2,7.2]$ and from 33.3\% to 2.5\% for $[3.7,4.7]$. Similar reductions are observed on LiveCodeBench, where disagreement decreases from 93.7\% to 28.6\% and from 86.9\% to 24.6\%, respectively.

Explicit Max also generally achieves higher maximum-score usage than Allowed List. On MATH-500, maximum-score usage increases from 72.4\% to 95.4\% for $[6.2,7.2]$ and from 94.3\% to 96.5\% for $[3.7,4.7]$. LiveCodeBench shows a similar pattern, with maximum-score usage increasing from 11.4\% to 65.7\% and from 42.3\% to 64.6\%, respectively. These results suggest that explicitly linking the upper scoring boundary to response quality can reduce evaluation instability associated with numerical-anchor misalignment.

\section{Conclusion}
In this paper, we introduce \textit{numerical scoring topology} as a structural factor in LLM-as-a-Judge evaluation and systematically examine its effects on zero-shot judgments. Our results show that numerical scoring topology affects LLM-as-a-Judge evaluation. Judgment-intensive tasks are particularly sensitive, while evaluation language can substantially modulate topology sensitivity, producing larger or qualitatively different effects across language and task conditions. Specific boundary configurations also cause greater instability than score width alone. Stronger and larger judge models are generally more robust to these changes. Furthermore, prompt-based mitigation strategies, particularly explicit maximum-score instructions, can reduce evaluation instability. These findings highlight the need to consider and standardize numerical scoring schemes in LLM-based evaluation.

\section*{Limitations}
This work has several limitations that also suggest directions for future research. First, although our experiments cover diverse evaluation benchmarks, scoring topologies, and seventeen LLM judges from both proprietary and open-source model families, they cannot encompass the full diversity of rapidly evolving foundation models. Evaluating future generations of LLM judges will help assess the generality of the observed phenomena.

Second, we focus on point-based numerical scoring. It remains an open question whether similar sensitivity to numerical scoring topology arises in alternative evaluation protocols, such as pairwise preference, ranking-based evaluation, or rubric-guided assessment.

Third, our multilingual analysis considers English and Korean as response languages and English, Korean, Chinese, and Japanese as evaluation languages. Although these settings cover multiple cross-lingual conditions, they do not represent the full diversity of world languages.

Finally, although our results provide consistent evidence that numerical scoring topology influences LLM-as-a-Judge evaluation, the mechanisms underlying these effects remain unclear. Understanding why specific boundary configurations alter evaluator behavior and incorporating these insights into more principled evaluation protocols remain important directions for future research.

\section*{Ethical Considerations}
This work uses existing benchmarks and model-generated responses and does not involve human participants or the collection of personal data. However, numerical scoring topology can influence evaluator outcomes and model comparisons, potentially leading to misleading conclusions when scoring schemes are treated as interchangeable. We therefore encourage researchers to report scoring ranges and normalization procedures transparently and to interpret LLM-judge scores with appropriate caution.

\section*{Acknowledgement}
This work was supported by Korea Internet \& Security Agency(KISA) grant funded by the Korea government(PIPC) (No.RS-2026-25526342, Development of Technologies for Preventing Sensitive Information Inference and Risk Assessment in Foundation Model Operations). This work was also supported by the National Research Foundation of Korea(NRF) grant funded by the Korea government(MSIT) (RS-2026-25483747). In addition, this research was supported by Culture, Sports and Tourism R\&D Program through the Korea Creative Content Agency grant funded by the Ministry of Culture, Sports and Tourism in 2026 (Project Name: Develop AI agent technology to connect knowledge through public cultural facility-based discussion and communication, Project Number: RS-2026-25520645). Furthermore, this work was supported by Institute of Information \& communications Technology Planning \& Evaluation (IITP, AI Computing Support Project for R\&D) grant funded by the Korea government(MSIT) (High-Performance Research AI Computing Infrastructure Support at the 2 PFLOPS Scale, RS-2026-25505492). Lastly, we gratefully acknowledge VESSL for providing GPU resources.

\bibliography{custom}

\clearpage

\appendix

\section{Experimental Setup Details}
\label{appendix:experimental-setup}

\subsection{Scoring Topologies}
We compare thirteen numerical scoring topologies that include both commonly used evaluation scales and deliberately constructed numerical variants. We organize them into four categories.

\begin{enumerate}

\item \textbf{Standard Scales}: Unit-interval (0--1), Likert-5 (0--5), and Likert-10 (0--10). These scales represent conventional scoring formats commonly used in LLM evaluation.

\item \textbf{Fine-grained Linear Scales}: Linear-50 (0--50), Linear-100 (0--100), and Linear-1000 (0--1000). These scales increase the numerical range and available score resolution while retaining conventional boundaries that begin at zero.

\item \textbf{Shifted and Narrow Ranges}: Positive Offset (3--15), Narrow Range (0.3--0.7), Symmetric Range ($-5$--5), Asymmetric Range ($-5$--2), and Shifted Range (0.74--1.48). These topologies modify the width, offset, symmetry, or location of the scoring range.

\item \textbf{Irregular Boundary Scales}: Irregular Range I ($-3.7$--8.9) and Irregular Range II ($-2.35$--7.85). These scales use nonstandard decimal boundaries that differ from conventional integer-based scoring ranges.

\end{enumerate}

In addition to these thirteen scoring topologies, we conduct controlled boundary experiments, shown in Table~\ref{tab:ablation_full}, to separate the effects of score width, boundary placement, and boundary alignment. These experiments address RQ2 by identifying which numerical properties are most strongly associated with evaluator instability.

\subsection{Evaluation Metrics}
Because each scoring topology uses a different numerical range, raw scores are not directly comparable. We therefore normalize each raw judge score $S_{\mathrm{raw}}$ to a common scale from 0 to 100.

% Because each scoring topology operates over a different numerical range, raw scores are not directly comparable. We therefore normalize every raw judge score $S_{\mathrm{raw}}$ into a common percentage scale:

\begin{equation}
S_{\mathrm{norm}}
=
\frac{S_{\mathrm{raw}}-S_{\min}}
{S_{\max}-S_{\min}}
\times100.
\end{equation}

where $S_{\min}$ and $S_{\max}$ denote the minimum and maximum values of the corresponding scoring topology. This transformation preserves the relative position of each score within its original range and allows comparison across different scoring schemes. 

To measure sensitivity to numerical scoring topology, we use the \textbf{Disagreement Rate}. For a scoring topology $t$, we compute the proportion of matched evaluation instances whose normalized scores differ from those obtained under the Likert-5 topology by more than a threshold $\delta$:

\begin{equation}
\mathrm{DR}_{\delta}(t)
=
\frac{1}{N}
\sum_{i=1}^{N}
\mathbb{I}
\left(
\left|
S_{i,t}^{\mathrm{norm}}
-
S_{i,\mathrm{Likert5}}^{\mathrm{norm}}
\right|
>
\delta
\right),
\end{equation}

where each matched evaluation instance $i$ corresponds to the same benchmark example, model response, evaluation prompt, and judge model under topology $t$ and the Likert-5 reference. $N$ denotes the total number of matched instances, and $\delta \in \{10,20\}$. We use Likert-5 as the reference because it is a conventional scoring scheme in LLM-as-a-Judge evaluation. A higher disagreement rate indicates greater sensitivity to numerical scoring topology.

For benchmarks with binary correctness labels, we measure the discriminative validity of each judge using the Area Under the Receiver Operating Characteristic Curve (AUROC)~\cite{airola2011experimental}. AUROC represents the probability that a randomly selected correct response receives a higher judge score than a randomly selected incorrect response, with ties receiving half credit:

\begin{equation}
\begin{aligned}
\mathrm{AUROC}
&=
P\left(
s_{\mathrm{correct}}
>
s_{\mathrm{incorrect}}
\right) \\
&\quad+
\frac{1}{2}
P\left(
s_{\mathrm{correct}}
=
s_{\mathrm{incorrect}}
\right).
\end{aligned}
\end{equation}

where $s_{\mathrm{correct}}$ and $s_{\mathrm{incorrect}}$ denote the judge scores assigned to correct and incorrect responses, respectively. An AUROC of 0.5 indicates random discrimination, while a higher AUROC indicates that the judge more reliably assigns higher scores to correct responses.

\subsection{Inference Settings}
\label{appendix:inference-settings}

We set the temperature to 0.0 for all judge models to minimize sampling variability. The maximum output length is set to 1,024 tokens by default and increased to 8,192--16,384 tokens for reasoning-oriented models, including GPT-OSS, DeepSeek, Qwen3, and MiniMax, to prevent truncation caused by reasoning-token consumption.

For GPT-5-mini and GPT-5.4, we use the low reasoning-effort setting. For all other models, the reasoning configuration is left unspecified, and the provider default is used.

Except for the prompt-based mitigation experiments in RQ5, all evaluations use the scale-only prompt configuration, which instructs the judge to return a single numerical score within the specified range without rubric-specific numerical anchors. We do not use the rubric-aware prompt variant. These settings are held fixed across scoring topologies for each judge model.

\section{Prompts}
\label{appendix:prompts}

This section presents the prompts used for response generation and evaluation. Placeholders enclosed in braces are replaced at runtime. The reference-answer field is included only when a reference answer is available.

\subsection{Generation Prompts}
\label{appendix:generation-prompts}

The generation prompts instruct the model to solve each benchmark problem carefully, provide a clear reasoning process, and conclude with a final answer. The prompts are designed to keep responses self-contained and differ only in language, with equivalent English and Korean versions used to control for linguistic effects during response generation. The complete English and Korean generation prompts are provided in Box~\ref{appendix:english-generation-prompt} and Box~\ref{appendix:korean-generation-prompt}, respectively.

\subsection{Judge Prompts}
\label{appendix:judge-prompts}

The judge prompts instruct the evaluator to assess each candidate response in terms of correctness, completeness, and overall quality. The judge must return a single numerical score within the specified range without additional explanation. The question, candidate response, and, when available, reference answer are included in the prompt. Equivalent English, Korean, Chinese, and Japanese versions are used in the corresponding language-condition experiments to isolate the effect of evaluation language while keeping the evaluation criteria unchanged. The complete English, Korean, Chinese, and Japanese judge prompts are provided in Box~\ref{appendix:english-judge-prompt}, Box~\ref{appendix:korean-judge-prompt}, Box~\ref{appendix:chinese-judge-prompt}, and Box~\ref{appendix:japanese-judge-prompt}, respectively.

\subsection{Prompt-Based Mitigation Prompts}
\label{appendix:mitigation-prompts}

The prompt-based mitigation strategies aim to reduce evaluation instability induced by numerical scoring topology. We consider two strategies, Allowed List and Explicit Max, while retaining the same evaluation criteria as the scale-only baseline. The Allowed List prompt explicitly provides a list of valid score values and instructs the judge to select one of them. The placeholder \texttt{\{allowed\_scores\}} is replaced with the corresponding score list at runtime. The Explicit Max prompt instructs the judge to assign the specified maximum score to a fully correct answer, the minimum score to a completely incorrect answer, and intermediate scores to partially correct answers. The complete Allowed List and Explicit Max prompts are provided in Box~\ref{appendix:allowed-list-prompt} and Box~\ref{appendix:explicit-max-prompt}, respectively.

\section{Additional Results}
\label{appendix:additional-results}

This appendix shows additional results that complement the main findings. We provide supplementary analyses on open-ended evaluation benchmarks, boundary-level properties of numerical scoring ranges, response--evaluation language conditions, and judge-level robustness. Together, these results provide a more detailed view of how numerical scoring topology affects both the absolute scores assigned by LLM judges and the stability of those scores across evaluation settings.

Table~\ref{tab:appendix-open-ended-scores} presents the results on WildBench and WritingBench. Despite normalization to a common 0--100 scale, evaluation scores vary substantially across scoring topologies. The symmetric and asymmetric ranges produce the highest scores on both benchmarks, exceeding the Likert-5 reference by approximately 1--2 points. In contrast, the narrow and shifted ranges consistently produce lower scores. In particular, the shifted range shows the largest decrease, with reductions of 9.3 points on WildBench and 10.1 points on WritingBench. These results further demonstrate that normalization does not eliminate the influence of the original scoring-range structure on judge behavior.

Table~\ref{tab:appendix-open-ended-disagreement} shows disagreement rates relative to the Likert-5 reference on WildBench and WritingBench. The shifted range produces the highest disagreement on both benchmarks, with 41.4\% of WildBench instances and 42.7\% of WritingBench instances differing by more than 10 normalized points. The narrow range also shows consistently high disagreement, particularly on WritingBench. In contrast, conventional integer-valued ranges, such as Likert-10 and Linear-50, produce substantially lower disagreement. These results indicate that narrow and shifted scoring ranges are more sensitive to numerical scoring topology than standard integer-aligned ranges, even on judgment-intensive benchmarks.

Table~\ref{tab:appendix-ablation-score-concentration} provides a detailed analysis of the boundary-ablation experiments by showing the most frequently assigned raw score under each scoring-range configuration. Across the width-sweep conditions, the most frequent score consistently coincides with the specified maximum, although the degree of concentration varies with range width. The round-maximum conditions exhibit particularly strong upper-bound concentration, with 77.0--82.1\% of evaluations assigned the maximum score. In contrast, two non-round maximum conditions concentrate on nearby salient values below the specified upper boundary: 4.6 instead of 4.7 and 7.0 instead of 7.2. However, this misalignment is not universal, as the remaining non-round and narrow non-round settings retain concentration at the true maximum. These patterns suggest that judge outputs depend on the interaction between boundary placement and numerical salience rather than on range width or decimal-valued boundaries alone.

\begin{figure*}[t]

    \subsection{Raw-Score Concentration and Boundary Misalignment}
    \label{appendix:raw-score-concentration}

    \centering
    \includegraphics[width=\textwidth]{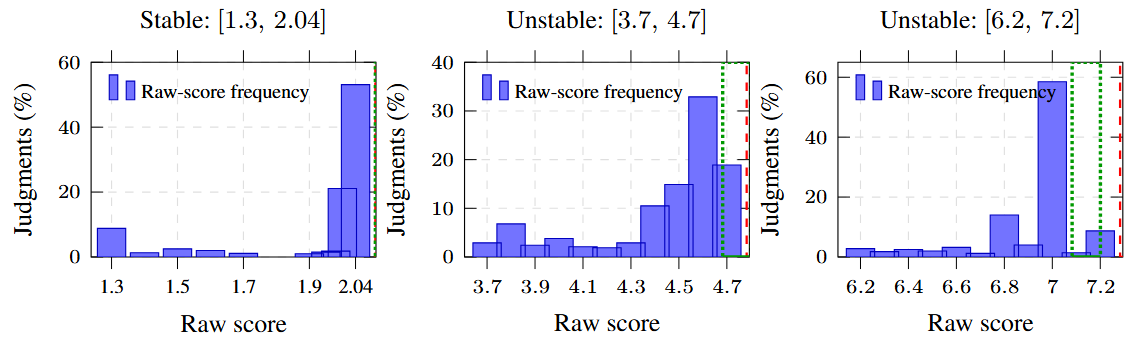}
    \caption{
    \textbf{Raw-score distributions under representative stable and unstable
    numerical scoring ranges using GPT-5-mini.}
    In the stable range $[1.3,2.04]$, the most frequent score coincides with the
    specified maximum, while an additional concentration at 2.00 remains close to
    the upper boundary after normalization. In contrast, the unstable ranges
    $[3.7,4.7]$ and $[6.2,7.2]$ concentrate on salient values below their
    specified maxima. In particular, 7.0 accounts for 58.5\% of judgments in
    $[6.2,7.2]$ but maps to only 80 on the normalized 0--100 scale. The dashed red
    and dotted green lines denote the specified maximum and the most frequent
    score, respectively.
    }
    \label{fig:raw-score-histograms}
\end{figure*}

Figure~\ref{fig:raw-score-histograms} examines the raw-score distributions under representative stable and unstable numerical scoring ranges. In the stable range $[1.3,2.04]$, the most frequent score coincides with the specified maximum, indicating strong alignment between the judge's preferred score and the upper boundary. Although an additional concentration appears at 2.00, this value remains close to the maximum after normalization.

In contrast, the unstable ranges $[3.7,4.7]$ and $[6.2,7.2]$ concentrate on salient values below their specified maxima. The range $[3.7,4.7]$ peaks at 4.6 rather than 4.7, while $[6.2,7.2]$ shows a particularly strong concentration at 7.0. Although 7.0 is numerically close to the upper boundary, it corresponds to only 80 on the normalized 0--100 scale, producing a large difference from the Likert-5 reference. These distributions provide direct evidence that instability arises from misalignment between the specified boundary and the score values preferred by the judge.

Table~\ref{tab:language-condition-disagreement} examines whether these effects persist across response--evaluation language conditions. The shifted range produces the highest disagreement in most English--Korean and English--Chinese conditions, with particularly large increases on Arena-Hard and IFEval. However, the exact pattern is language dependent: under Japanese evaluation, the narrow range shows the largest disagreement on several benchmarks, while the shifted range remains highly unstable. Overall, the shifted and narrow ranges are consistently among the most sensitive scoring topologies, although their relative ordering varies across tasks and evaluation languages. These results show that sensitivity to numerical score design extends beyond a single response or evaluation language.

Table~\ref{tab:appendix-score-gap-mad} presents supplementary results on score separation and cross-topology deviation across judge models. GPT-OSS-120B achieves the largest score gap between correct and incorrect responses, while GPT-5.4 and Qwen3.7-Plus exhibit the lowest mean absolute deviation between Likert-5 and shifted-range scores. Several smaller judge models display substantially greater sensitivity to scoring topology. In particular, Llama-3.1-8B and Gemma-3-4B show mean absolute deviations above 50 points, indicating that their evaluations change considerably when only the numerical score range is modified. Overall, stronger judge models tend to combine better separation between correct and incorrect responses with greater robustness to numerical scoring topology, although the relationship is not uniform across all models.

Table~\ref{tab:response-generator-scores} presents the mean normalized judge scores of eight response generators across six scoring topologies on MATH-500 and Arena-Hard. On MATH-500, score differences across topologies are generally modest, although the Shifted and Narrow Ranges consistently assign lower scores than Likert-5. In contrast, Arena-Hard exhibits much larger topology-dependent variation. The Shifted and Narrow Ranges reduce scores across all generators, with larger decreases observed for several lower-ranked generators. Despite these changes in absolute scores, the relative ordering of the eight response generators remains unchanged under all evaluated topologies on both benchmarks.

Finally, Table~\ref{tab:response-generator-score-shifts} summarizes these differences as the average score change relative to Likert-5. On MATH-500, the mean shifts
range from $-2.7$ to $+0.6$ points, indicating relatively limited sensitivity. Arena-Hard shows substantially larger downward shifts, with the Shifted and
Narrow Ranges reducing the average score by 9.8 and 9.4 points, respectively. Nevertheless, the complete response-generator ranking is preserved, with
Kendall's $\tau=1.00$ and no top-1 or pairwise ranking flips. These results show that score stability and ranking stability are distinct: numerical
scoring topology can substantially alter absolute evaluation scores without changing the relative ordering of response generators.

\clearpage
\onecolumn

% ==================================================
% Generation prompt boxes: B.1.1, B.1.2
% ==================================================
\begingroup
\setcounter{subsubsection}{0}
\renewcommand{\thesubsubsection}{B.1.\arabic{subsubsection}}

\subsubsection{English Generation Prompt}
\label{appendix:english-generation-prompt}

\vspace{4pt}

\begin{englishpromptbox}{Box B.1.1: English Generation Prompt}
\begin{Verbatim}[fontsize=\small,breaklines=true,breakanywhere=true]
Solve the following problem carefully.

Show your reasoning clearly and end with a final answer.

Keep the response self-contained.

Problem:

{question}
\end{Verbatim}
\end{englishpromptbox}

\vspace{4pt}

\subsubsection{Korean Generation Prompt}
\label{appendix:korean-generation-prompt}

\vspace{4pt}

\begin{koreanpromptbox}{Box B.1.2: Korean Generation Prompt}
\begin{Verbatim}[fontsize=\small,breaklines=true,breakanywhere=true]
다음 문제를 신중하게 풀이하세요.

추론 과정을 명확히 보이고 마지막에 최종 답을 제시하세요.

응답은 그 자체로 완결되게 작성하세요.

문제:

{question}
\end{Verbatim}
\end{koreanpromptbox}

\endgroup

% ==================================================
% Judge prompt boxes: B.2.1, B.2.2
% ==================================================
\begingroup
\setcounter{subsubsection}{0}
\renewcommand{\thesubsubsection}{B.2.\arabic{subsubsection}}

\vspace{4pt}

\subsubsection{English Judge Prompt}
\label{appendix:english-judge-prompt}

\vspace{4pt}

\begin{englishpromptbox}{Box B.2.1: English Judge Prompt}
\begin{Verbatim}[fontsize=\small,breaklines=true,breakanywhere=true]
You are an impartial evaluation judge.

Evaluate the candidate answer for correctness, completeness, and quality.

Return only one numeric score between {min_score} and {max_score}.

Do not include explanations or any extra text.

Question:

{question}

[Included only when a reference answer is available]

Reference Answer:

{reference_answer}

Candidate Answer:

{candidate_answer}
\end{Verbatim}
\end{englishpromptbox}

\vspace{4pt}

\clearpage

\subsubsection{Korean Judge Prompt}
\label{appendix:korean-judge-prompt}

\vspace{4pt}

\begin{koreanpromptbox}{Box B.2.2: Korean Judge Prompt}
\begin{Verbatim}[fontsize=\small,breaklines=true,breakanywhere=true]
당신은 공정한 평가 심사자입니다.

후보 답변의 정확성, 완전성, 품질을 평가하세요.

{min_score}에서 {max_score} 사이의 숫자 점수 하나만 반환하세요.

설명이나 그 외 어떤 텍스트도 포함하지 마세요.

질문:

{question}

[참고 답안이 있는 경우에만 포함]

참고 답안:

{reference_answer}

후보 답변:

{candidate_answer}
\end{Verbatim}
\end{koreanpromptbox}

\endgroup

% ==================================================
% Additional judge prompt boxes: B.2.3, B.2.4
% ==================================================
\begingroup
\setcounter{subsubsection}{2}
\renewcommand{\thesubsubsection}{B.2.\arabic{subsubsection}}

\vspace{4pt}

\subsubsection{Chinese Judge Prompt}
\label{appendix:chinese-judge-prompt}

\vspace{4pt}

\begin{koreanpromptbox}{Box B.2.3: Chinese Judge Prompt}
\begin{CJK*}{UTF8}{gbsn}
\small
\setlength{\parindent}{0pt}

你是一位公正的评估评审。

\medskip
请评估候选答案的正确性、完整性和质量。

\medskip
只返回一个介于 \texttt{\{min\_score\}} 和
\texttt{\{max\_score\}} 之间的数字分数。

\medskip
不要包含任何解释或额外的文字。

\medskip
问题：

\medskip
\texttt{\{question\}}

\medskip
[仅在有参考答案时包含]

\medskip
参考答案：

\medskip
\texttt{\{reference\_answer\}}

\medskip
候选答案：

\medskip
\texttt{\{candidate\_answer\}}
\end{CJK*}
\end{koreanpromptbox}

% \begin{koreanpromptbox}{Box B.2.3: Chinese Judge Prompt}
% \begin{Verbatim}[fontsize=\small,breaklines=true,breakanywhere=true]

% 你是一位公正的评估评审。

% 请评估候选答案的正确性、完整性和质量。

% 只返回一个介于 {min_score} 和 {max_score} 之间的数字分数。

% 不要包含任何解释或额外的文字。

% 问题：

% {question}

% [仅在有参考答案时包含]

% 参考答案：

% {reference_answer}

% 候选答案：

% {candidate_answer}
% \end{Verbatim}
% \end{koreanpromptbox}

\vspace{4pt}

\clearpage

\subsubsection{Japanese Judge Prompt}
\label{appendix:japanese-judge-prompt}

\vspace{4pt}

\begin{koreanpromptbox}{Box B.2.4: Japanese Judge Prompt}
\begin{CJK*}{UTF8}{min}
\small
\setlength{\parindent}{0pt}

あなたは公正な評価者です。

\medskip
候補回答の正確性、完全性、品質を評価してください。

\medskip
\texttt{\{min\_score\}} から \texttt{\{max\_score\}} の間の
数値スコアを一つだけ返してください。

\medskip
説明やその他のテキストは一切含めないでください。

\medskip
質問：

\medskip
\texttt{\{question\}}

\medskip
[参考回答がある場合のみ含める]

\medskip
参考回答：

\medskip
\texttt{\{reference\_answer\}}

\medskip
候補回答：

\medskip
\texttt{\{candidate\_answer\}}
\end{CJK*}
\end{koreanpromptbox}

% \begin{koreanpromptbox}{Box B.2.4: Japanese Judge Prompt}
% \begin{Verbatim}[fontsize=\small,breaklines=true,breakanywhere=true]
% あなたは公正な評価者です。

% 候補回答の正確性、完全性、品質を評価してください。

% {min_score} から {max_score} の間の数値スコアを一つだけ返してください。

% 説明やその他のテキストは一切含めないでください。

% 質問：

% {question}

% [参考回答がある場合のみ含める]

% 参考回答：

% {reference_answer}

% 候補回答：

% {candidate_answer}
% \end{Verbatim}
% \end{koreanpromptbox}

\endgroup

\clearpage

% ==================================================
% Prompt-based mitigation boxes: B.3.1, B.3.2
% ==================================================
\begingroup
\setcounter{subsubsection}{0}
\renewcommand{\thesubsubsection}{B.3.\arabic{subsubsection}}

\vspace{4pt}

\subsubsection{Allowed List Prompt}
\label{appendix:allowed-list-prompt}

\vspace{4pt}

\begin{englishpromptbox}{Box B.3.1: Allowed List Prompt}
\begin{Verbatim}[fontsize=\small,breaklines=true,breakanywhere=true]
You are an impartial evaluation judge.

Evaluate the candidate answer for correctness, completeness, and quality.

Return only one numeric score between {min} and {max}.

Do not include explanations or any extra text.

Choose exactly one score from this list of allowed values:
{allowed_scores}

Question:
{question}

Reference Answer:
{reference_answer}

Candidate Answer:
{candidate_answer}
\end{Verbatim}
\end{englishpromptbox}

\vspace{4pt}

\subsubsection{Explicit Max Prompt}
\label{appendix:explicit-max-prompt}

\vspace{4pt}

\begin{englishpromptbox}{Box B.3.2: Explicit Max Prompt}
\begin{Verbatim}[fontsize=\small,breaklines=true,breakanywhere=true]
You are an impartial evaluation judge.

Evaluate the candidate answer for correctness, completeness, and quality.

Return only one numeric score between {min} and {max}.

Do not include explanations or any extra text.

Use {max} for a fully correct, perfect answer and {min} for a completely incorrect answer; 
use intermediate values for partially correct answers.

Question:
{question}

Reference Answer:
{reference_answer}

Candidate Answer:
{candidate_answer}
\end{Verbatim}
\end{englishpromptbox}

\endgroup

\clearpage
\twocolumn

\begin{table*}[t]

\subsection{Performance on Judgment-Intensive Benchmarks}
\label{appendix:open-ended-scores}

\vspace{6pt}

\centering
\scriptsize
\setlength{\tabcolsep}{10pt}
\renewcommand{\arraystretch}{1.05}
\begin{tabular}{lcc}
\toprule
\textbf{Scoring Scale}
& \textbf{WildBench}
& \textbf{WritingBench} \\
\midrule
Unit-interval (0 to 1)
& \vn{81.3}{4.0}
& \vn{86.0}{3.9} \\

\textbf{Likert-5 (0 to 5)}
& \textbf{\vz{85.3}}
& \textbf{\vz{89.9}} \\

Likert-10 (0 to 10)
& \vn{81.9}{3.4}
& \vn{84.5}{5.4} \\

Linear-50 (0 to 50)
& \vn{82.4}{2.9}
& \vn{87.1}{2.8} \\

Linear-100 (0 to 100)
& \vn{83.4}{1.9}
& \vn{85.9}{4.1} \\

Linear-1000 (0 to 1000)
& \vn{82.1}{3.3}
& \vn{86.3}{3.6} \\

Positive Offset (3 to 15)
& \vn{81.1}{4.2}
& \vn{85.6}{4.3} \\

Narrow Range (0.3 to 0.7)
& \vn{77.3}{8.0}
& \vn{79.6}{10.3} \\

Symmetric Range ($-5$ to 5)
& \vp{86.9}{1.6}
& \vp{91.4}{1.4} \\

Asymmetric Range ($-5$ to 2)
& \vp{86.6}{1.3}
& \vp{91.3}{1.4} \\

Shifted Range (0.74 to 1.48)
& \vn{76.0}{9.3}
& \vn{79.9}{10.1} \\

Irregular Range I ($-3.7$ to 8.9)
& \vn{84.3}{1.0}
& \vn{89.5}{0.5} \\

Irregular Range II ($-2.35$ to 7.85)
& \vp{85.4}{0.1}
& \vn{88.9}{1.0} \\
\bottomrule
\end{tabular}

\caption{\textbf{Performance on WildBench and WritingBench across numerical scoring topologies.}
GPT-5-mini is used both to generate responses and to evaluate them under all scoring topologies. All scores are normalized to a 0--100 scale. Values in parentheses denote deviations from the Likert-5 reference.}
\label{tab:appendix-open-ended-scores}
\end{table*}

\begin{table*}[t]

\subsection{Judgment-Intensive Benchmark Disagreement}
\label{appendix:open-ended-disagreement}

\vspace{6pt}

\centering
\scriptsize
\setlength{\tabcolsep}{8pt}
\renewcommand{\arraystretch}{1.05}
\begin{tabular}{lcccc}
\toprule
\multirow{2}{*}{\textbf{Scoring Scale}}
& \multicolumn{2}{c}{\textbf{WildBench}}
& \multicolumn{2}{c}{\textbf{WritingBench}} \\
\cmidrule(lr){2-3}
\cmidrule(lr){4-5}
&
$\mathbf{|\Delta|>10}$
&
$\mathbf{|\Delta|>20}$
&
$\mathbf{|\Delta|>10}$
&
$\mathbf{|\Delta|>20}$ \\
\midrule
Unit-interval (0 to 1)
& 18.8 (48) & 8.6 (22)
& 18.7 (104) & 6.7 (37) \\

Likert-10 (0 to 10)
& 13.7 (35) & 3.9 (10)
& 9.0 (50) & 1.4 (8) \\

Linear-50 (0 to 50)
& 16.8 (43) & 3.5 (9)
& 16.4 (91) & 3.1 (17) \\

Linear-100 (0 to 100)
& 17.2 (44) & 4.7 (12)
& 14.1 (78) & 3.8 (21) \\

Linear-1000 (0 to 1000)
& 18.0 (46) & 9.0 (23)
& 15.7 (87) & 4.3 (24) \\

Positive Offset (3 to 15)
& 22.3 (57) & 9.8 (25)
& 21.4 (119) & 5.2 (29) \\

Narrow Range (0.3 to 0.7)
& 34.8 (89) & \textbf{14.8} (38)
& 38.4 (213) & 13.0 (72) \\

Symmetric Range ($-5$ to 5)
& 19.1 (49) & 7.0 (18)
& 14.1 (78) & 4.3 (24) \\

Asymmetric Range ($-5$ to 2)
& 24.6 (63) & 10.9 (28)
& 20.0 (111) & 8.5 (47) \\

Shifted Range (0.74 to 1.48)
& \textbf{41.4} (106) & 13.7 (35)
& \textbf{42.7} (237) & \textbf{15.5} (86) \\

Irregular Range I ($-3.7$ to 8.9)
& 25.0 (64) & 8.2 (21)
& 19.6 (109) & 4.5 (25) \\

Irregular Range II ($-2.35$ to 7.85)
& 25.8 (66) & 7.8 (20)
& 18.2 (101) & 4.5 (25) \\
\bottomrule
\end{tabular}

\caption{\textbf{Disagreement rates on WildBench and WritingBench across numerical scoring topologies using GPT-5-mini.}
Values report the percentage of instances whose normalized scores differ from the Likert-5 reference by more than 10 or 20 points, with counts shown in parentheses. Higher values indicate greater sensitivity to numerical scoring topology.}
\label{tab:appendix-open-ended-disagreement}
\end{table*}

\begin{table*}[t]

\subsection{Boundary-Ablation Results}
\label{appendix:boundary-ablation-results}

\vspace{6pt}

\centering
\small
\setlength{\tabcolsep}{10pt}
\renewcommand{\arraystretch}{1.08}
\begin{tabular}{llccc}
\toprule
\textbf{Factor}
& \textbf{Scoring Range}
& \textbf{MFS}
& \textbf{Count}
& \textbf{Prop.} \\
\midrule

\multirow{4}{*}{Width Sweep}
& 0.74--2.48
& 2.48
& 993
& 43.5\% \\

& 0.74--4.74
& 4.74
& 751
& 32.9\% \\

& 0.74--8.74
& 8.74
& 584
& 25.6\% \\

& 0.74--15.74
& 15.74
& 959
& 42.0\% \\

\midrule

\multirow{3}{*}{Round Max}
& 1.0--2.0
& 2.0
& 1,876
& 82.1\% \\

& 3.0--4.0
& 4.0
& 1,759
& 77.0\% \\

& 6.0--7.0
& 7.0
& 1,780
& 77.9\% \\

\midrule

\multirow{3}{*}{Non-round Max}
& 1.3--2.04
& 2.04
& 1,213
& 53.1\% \\

& \textbf{3.7--4.7}
& \textbf{4.6}
& \textbf{752}
& \textbf{32.9\%} \\

& \textbf{6.2--7.2}
& \textbf{7.0}
& \textbf{1,335}
& \textbf{58.5\%} \\

\midrule

\multirow{2}{*}{Narrow Non-round}
& 2.11--2.85
& 2.85
& 1,001
& 43.8\% \\

& 3.33--4.07
& 4.07
& 888
& 38.9\% \\

\bottomrule
\end{tabular}

\caption{\textbf{Most frequently assigned raw scores across boundary-ablation settings using GPT-5-mini.}
MFS denotes the most frequently assigned score, and Prop. denotes the proportion of evaluations assigned that score. Results are aggregated across 2,284 instances from MATH-500, AIME, GPQA Main, Arena-Hard, LiveCodeBench, and IFEval. Bold rows indicate the two boundary configurations with the highest disagreement rates.}
\label{tab:appendix-ablation-score-concentration}
\end{table*}

\clearpage
\onecolumn

\subsection{Response--Evaluation Language Results}
\label{appendix:language-condition-results}

\begingroup
\scriptsize
\setlength{\tabcolsep}{3.2pt}
\renewcommand{\arraystretch}{1.05}

\begin{longtable}{@{}lcccccccc@{}}

\toprule
\multirow{2}{*}{\textbf{Scoring Scale}}
& \multicolumn{2}{c}{\textbf{MATH-500}}
& \multicolumn{2}{c}{\textbf{LCB}}
& \multicolumn{2}{c}{\textbf{Arena-Hard}}
& \multicolumn{2}{c}{\textbf{IFEval}} \\
\cmidrule(lr){2-3}
\cmidrule(lr){4-5}
\cmidrule(lr){6-7}
\cmidrule(lr){8-9}
& $>10$ & $>20$
& $>10$ & $>20$
& $>10$ & $>20$
& $>10$ & $>20$ \\
\midrule
\endfirsthead

\multicolumn{9}{c}{\tablename\ \thetable\ continued} \\
\addlinespace[2pt]
\toprule
\multirow{2}{*}{\textbf{Scoring Scale}}
& \multicolumn{2}{c}{\textbf{MATH-500}}
& \multicolumn{2}{c}{\textbf{LCB}}
& \multicolumn{2}{c}{\textbf{Arena-Hard}}
& \multicolumn{2}{c}{\textbf{IFEval}} \\
\cmidrule(lr){2-3}
\cmidrule(lr){4-5}
\cmidrule(lr){6-7}
\cmidrule(lr){8-9}
& $>10$ & $>20$
& $>10$ & $>20$
& $>10$ & $>20$
& $>10$ & $>20$ \\
\midrule
\endhead

\multicolumn{9}{r}{\textit{Continued on next page}} \\
\endfoot

\endlastfoot

\multicolumn{9}{@{}l}{\textbf{EN$\rightarrow$KO}} \\
\addlinespace[2pt]

Unit-interval (0 to 1)
& 11.6 & 4.4
& 16.6 & 8.0
& 6.8 & 2.6
& 32.5 & 19.6 \\

Likert-10 (0 to 10)
& 6.2 & 3.6
& 10.9 & 6.3
& 5.6 & 2.0
& 27.9 & 15.5 \\

Linear-50 (0 to 50)
& 5.6 & 2.6
& 13.1 & 8.0
& 9.4 & 3.0
& 26.8 & 15.2 \\

Linear-100 (0 to 100)
& 7.0 & 3.4
& 14.3 & 8.6
& 8.2 & 2.6
& 28.1 & 15.0 \\

Linear-1000 (0 to 1000)
& 6.4 & 2.8
& 14.9 & 9.7
& 9.0 & 3.0
& 28.3 & 16.5 \\

Positive Offset (3 to 15)
& 10.4 & 4.6
& 12.6 & 6.9
& 11.2 & 4.4
& 33.6 & 17.7 \\

Narrow Range (0.3 to 0.7)
& 14.6 & 5.6
& 23.4 & 7.4
& 44.0 & 6.0
& 47.3 & 22.7 \\

Symmetric Range ($-5$ to 5)
& 7.6 & 3.6
& 12.0 & 8.0
& 7.0 & 2.2
& 29.0 & 18.3 \\

Asymmetric Range ($-5$ to 2)
& 8.8 & 4.4
& 16.6 & 6.9
& 8.4 & 2.4
& 32.3 & 19.4 \\

Shifted Range (0.74 to 1.48)
& \textbf{22.4} & \textbf{12.8}
& \textbf{47.4} & \textbf{21.7}
& \textbf{53.8} & \textbf{13.4}
& \textbf{53.6} & \textbf{29.4} \\

Irregular Range I ($-3.7$ to 8.9)
& 13.4 & 7.6
& 13.1 & 8.0
& 12.2 & 3.2
& 33.1 & 18.5 \\

Irregular Range II ($-2.35$ to 7.85)
& 13.2 & 7.6
& 16.0 & 10.9
& 11.6 & 2.2
& 34.2 & 19.2 \\

\midrule
\multicolumn{9}{@{}l}{\textbf{KO$\rightarrow$EN}} \\
\addlinespace[2pt]

Unit-interval (0 to 1)
& 2.4 & 0.6
& 13.1 & 5.7
& 4.8 & 1.8
& 32.0 & 15.3 \\

Likert-10 (0 to 10)
& 1.6 & 0.4
& 6.3 & 2.9
& 5.8 & 1.2
& 21.1 & 12.4 \\

Linear-50 (0 to 50)
& 0.4 & 0.2
& 7.4 & 3.4
& 5.4 & 1.8
& 22.6 & 12.4 \\

Linear-100 (0 to 100)
& 0.8 & 0.2
& 10.3 & 7.4
& 6.6 & 1.6
& 25.9 & 13.3 \\

Linear-1000 (0 to 1000)
& 1.0 & 0.4
& 11.4 & 6.3
& 7.4 & 2.2
& 23.7 & 14.4 \\

Positive Offset (3 to 15)
& 2.2 & 0.4
& 12.0 & 4.6
& 9.6 & 3.2
& 31.1 & 17.7 \\

Narrow Range (0.3 to 0.7)
& 2.8 & 1.4
& 22.3 & 9.7
& 22.2 & 5.6
& 37.3 & 20.1 \\

Symmetric Range ($-5$ to 5)
& 1.2 & 0.2
& 11.4 & 5.7
& 4.4 & 1.4
& 29.6 & 16.5 \\

Asymmetric Range ($-5$ to 2)
& 1.2 & 0.4
& 11.4 & 5.1
& 7.2 & 2.4
& 28.5 & 16.6 \\

Shifted Range (0.74 to 1.48)
& \textbf{3.6} & 0.2
& \textbf{24.0} & 5.1
& \textbf{25.0} & \textbf{6.0}
& \textbf{43.1} & 17.4 \\

Irregular Range I ($-3.7$ to 8.9)
& 2.6 & 1.0
& 14.3 & 5.1
& 8.0 & 3.2
& 30.9 & \textbf{20.3} \\

Irregular Range II ($-2.35$ to 7.85)
& 2.0 & 0.8
& 12.0 & 6.3
& 5.6 & 1.0
& 33.1 & 16.8 \\

\midrule
\multicolumn{9}{@{}l}{\textbf{KO$\rightarrow$KO}} \\
\addlinespace[2pt]

Unit-interval (0 to 1)
& 2.8 & 0.8
& 14.9 & 8.0
& 7.4 & 3.2
& 38.3 & 22.0 \\

Likert-10 (0 to 10)
& 1.8 & 0.6
& 8.6 & 6.3
& 5.4 & 1.8
& 30.1 & 18.5 \\

Linear-50 (0 to 50)
& 1.4 & 0.4
& 9.7 & 9.1
& 7.2 & 2.4
& 34.2 & 17.9 \\

Linear-100 (0 to 100)
& 1.8 & 0.4
& 7.4 & 4.6
& 7.8 & 2.6
& 32.3 & 19.4 \\

Linear-1000 (0 to 1000)
& 1.2 & 0.4
& 10.3 & 6.3
& 8.4 & 3.6
& 30.1 & 18.5 \\

Positive Offset (3 to 15)
& 3.8 & 2.6
& 11.4 & 6.3
& 9.2 & 4.2
& 38.3 & 22.9 \\

Narrow Range (0.3 to 0.7)
& 8.2 & 0.8
& 34.9 & 10.3
& 45.0 & 6.2
& 52.9 & 28.7 \\

Symmetric Range ($-5$ to 5)
& 1.8 & 0.8
& 8.6 & 5.7
& 5.2 & 2.2
& 33.5 & 21.4 \\

Asymmetric Range ($-5$ to 2)
& 2.0 & 0.8
& 10.3 & 5.7
& 9.2 & 4.2
& 37.2 & 23.8 \\

Shifted Range (0.74 to 1.48)
& \textbf{16.8} & \textbf{10.4}
& \textbf{37.1} & \textbf{20.0}
& \textbf{49.0} & \textbf{12.4}
& \textbf{62.3} & \textbf{30.5} \\

Irregular Range I ($-3.7$ to 8.9)
& 3.6 & 1.8
& 10.3 & 5.1
& 8.6 & 2.6
& 38.1 & 23.5 \\

Irregular Range II ($-2.35$ to 7.85)
& 3.8 & 2.2
& 14.9 & 9.1
& 10.4 & 2.8
& 40.9 & 23.8 \\

\midrule
\multicolumn{9}{@{}l}{\textbf{EN$\rightarrow$ZH}} \\
\addlinespace[2pt]

Unit-interval (0 to 1)
& 2.0 & 1.4
& 11.7 & 4.1
& 7.0 & 2.4
& 30.9 & 17.4 \\

Likert-10 (0 to 10)
& 2.6 & 1.6
& 11.4 & 6.9
& 6.8 & 3.2
& 25.7 & 15.3 \\

Linear-50 (0 to 50)
& 1.6 & 1.0
& 11.0 & 6.4
& 8.6 & 2.6
& 24.6 & 14.8 \\

Linear-100 (0 to 100)
& 1.2 & 1.0
& 12.0 & 8.6
& 8.6 & 2.8
& 29.2 & 17.3 \\

Linear-1000 (0 to 1000)
& 1.6 & 1.4
& 11.2 & 3.5
& 8.2 & 2.2
& 25.8 & 14.5 \\

Positive Offset (3 to 15)
& 2.2 & 1.6
& 15.2 & 9.4
& 10.4 & 3.8
& 35.4 & 20.8 \\

Narrow Range (0.3 to 0.7)
& 11.4 & 1.8
& 31.4 & 8.6
& 58.4 & 12.0
& 57.1 & 26.2 \\

Symmetric Range ($-5$ to 5)
& 1.4 & 1.0
& 9.9 & 7.6
& 6.4 & 2.6
& 26.0 & 15.8 \\

Asymmetric Range ($-5$ to 2)
& 2.0 & 1.2
& 15.0 & 7.5
& 10.8 & 3.8
& 30.7 & 20.3 \\

Shifted Range (0.74 to 1.48)
& \textbf{19.4} & \textbf{11.4}
& \textbf{44.6} & \textbf{19.4}
& \textbf{63.8} & \textbf{15.4}
& \textbf{57.0} & \textbf{23.9} \\

Irregular Range I ($-3.7$ to 8.9)
& 2.4 & 1.8
& 13.9 & 9.2
& 10.4 & 2.8
& 28.5 & 16.2 \\

Irregular Range II ($-2.35$ to 7.85)
& 2.6 & 1.8
& 13.7 & 9.1
& 12.2 & 3.6
& 30.6 & 15.9 \\

\midrule
\multicolumn{9}{@{}l}{\textbf{EN$\rightarrow$JA}} \\
\addlinespace[2pt]

Unit-interval (0 to 1)
& 2.2 & 1.4
& 11.6 & 7.0
& 7.8 & 3.2
& 31.8 & 18.4 \\

Likert-10 (0 to 10)
& 0.6 & 0.4
& 13.2 & 7.5
& 5.8 & 2.6
& 28.1 & 16.3 \\

Linear-50 (0 to 50)
& 1.2 & 0.8
& 10.4 & 8.7
& 8.8 & 3.2
& 30.5 & 17.0 \\

Linear-100 (0 to 100)
& 1.2 & 0.8
& 9.7 & 4.0
& 7.6 & 2.6
& 29.9 & 17.5 \\

Linear-1000 (0 to 1000)
& 1.0 & 1.0
& 13.4 & 8.1
& 9.3 & 2.6
& 28.7 & 16.7 \\

Positive Offset (3 to 15)
& 1.8 & 1.0
& 13.1 & 8.0
& 9.4 & 3.0
& 35.2 & 19.5 \\

Narrow Range (0.3 to 0.7)
& 4.6 & 1.0
& 26.3 & 12.6
& 47.8 & 8.6
& 49.1 & 25.9 \\

Symmetric Range ($-5$ to 5)
& 1.2 & 0.6
& 11.0 & 7.6
& 6.0 & 3.6
& 25.6 & 16.6 \\

Asymmetric Range ($-5$ to 2)
& 1.8 & 1.2
& 10.6 & 8.2
& 9.0 & 4.0
& 34.1 & 20.6 \\

Shifted Range (0.74 to 1.48)
& \textbf{5.6} & \textbf{3.0}
& \textbf{32.0} & \textbf{16.6}
& \textbf{31.8} & \textbf{5.8}
& \textbf{48.1} & \textbf{24.6} \\

Irregular Range I ($-3.7$ to 8.9)
& 2.0 & 1.4
& 16.2 & 9.8
& 9.2 & 2.8
& 34.1 & 18.1 \\

Irregular Range II ($-2.35$ to 7.85)
& 2.2 & 1.8
& 16.6 & 8.0
& 12.4 & 3.4
& 35.8 & 20.0 \\

\midrule

\caption{\textbf{Cross-topology disagreement rates across response--evaluation language conditions using GPT-5-mini.} Values report the percentage of instances whose normalized scores differ from the condition-specific Likert-5 reference by more than 10 or 20 points. Bold values indicate the highest disagreement rates within each benchmark and language condition.}
\label{tab:language-condition-disagreement}

\end{longtable}

\endgroup

\clearpage
\twocolumn

\begin{table*}[t]

\subsection{Judge-Level Robustness Results}
\label{appendix:judge-robustness-results}

\vspace{6pt}

\centering
\small
\setlength{\tabcolsep}{8pt}
\renewcommand{\arraystretch}{1.08}
\begin{tabular}{lcc}
\toprule
\textbf{Judge Model}
& \textbf{Score Gap}
& \textbf{MAD} \\
\midrule

GPT-5.4
& 75.1 & \textbf{3.7} \\

MiniMax-M3
& 71.5 & 7.2 \\

GPT-OSS-120B
& \textbf{82.9} & 8.3 \\

DeepSeek-V4-Pro
& 80.8 & 8.3 \\

GPT-5-mini
& 69.3 & 6.1 \\

Qwen3.7-Plus
& 80.8 & \textbf{3.7} \\

Gemini-2.5-Flash-Lite
& 55.1 & 13.4 \\

\midrule

GPT-4o-mini
& 44.7 & 7.1 \\

GPT-OSS-20B
& 79.4 & 10.3 \\

Llama-3.1-70B
& 23.9 & 41.3 \\

Llama-3.1-8B
& 11.2 & 56.5 \\

Gemma-3-27B
& 33.9 & 27.7 \\

Gemma-3-12B
& 40.6 & 26.1 \\

Gemma-3-4B
& 31.3 & 53.7 \\

Qwen3-32B
& 37.8 & 8.9 \\

Qwen3-14B
& 45.5 & 9.8 \\

Qwen3-8B
& 48.9 & 7.9 \\

\bottomrule
\end{tabular}

\caption{\textbf{Supplementary score separation and cross-topology deviation across judge models.}
Score Gap denotes the difference between the mean normalized scores assigned to correct and incorrect responses under Likert-5 scoring. MAD shows the mean absolute difference between normalized Likert-5 and Shifted-Range scores. Higher Score Gap indicates greater mean score separation, whereas lower MAD indicates greater robustness to changes in numerical scoring topology.}
\label{tab:appendix-score-gap-mad}
\end{table*}

\begin{table*}[t]

\subsection{Score Shifts and Ranking Stability across Response Generators}
\label{appendix:ranking-stability}

\vspace{6pt}

\centering
\small
\setlength{\tabcolsep}{4.5pt}
\renewcommand{\arraystretch}{1.06}

\begin{tabular}{lrrrrrr}
\toprule
\textbf{Response Generator}
& \textbf{Likert-5}
& \textbf{Likert-10}
& \textbf{Linear-100}
& \textbf{Shifted Range}
& \textbf{Narrow Range}
& \textbf{Irregular Range II} \\
\midrule

\multicolumn{7}{c}{\textbf{MATH-500}} \\
\midrule
GPT-5-mini
& 97.8 & 97.6 & 97.7 & 96.7 & 96.4 & 97.9 \\
GPT-OSS-120B
& 96.0 & 96.0 & 96.0 & 94.4 & 94.9 & 96.0 \\
GPT-OSS-20B
& 92.4 & 92.5 & 92.7 & 91.2 & 90.9 & 93.0 \\
Gemma-3-12B
& 85.9 & 85.9 & 86.2 & 83.3 & 83.5 & 85.8 \\
GPT-4o-mini
& 80.3 & 80.3 & 80.4 & 77.7 & 78.1 & 80.7 \\
Gemma-3-4B
& 76.8 & 76.4 & 77.4 & 73.3 & 74.7 & 78.0 \\
Llama-3.1-70B
& 66.9 & 67.3 & 67.2 & 62.8 & 64.5 & 68.5 \\
Llama-3.1-8B
& 52.2 & 51.7 & 53.2 & 47.1 & 48.6 & 53.0 \\

\midrule
\multicolumn{7}{c}{\textbf{Arena-Hard}} \\
\midrule
GPT-5-mini
& 96.4 & 92.2 & 92.5 & 89.2 & 87.1 & 94.5 \\
GPT-OSS-120B
& 91.1 & 87.6 & 89.5 & 84.3 & 83.9 & 90.1 \\
GPT-OSS-20B
& 83.6 & 79.7 & 83.2 & 75.1 & 75.0 & 83.9 \\
GPT-4o-mini
& 71.3 & 67.2 & 71.4 & 60.6 & 60.1 & 73.1 \\
Gemma-3-12B
& 69.3 & 65.4 & 68.5 & 59.3 & 58.4 & 68.7 \\
Llama-3.1-70B
& 58.6 & 56.4 & 59.6 & 46.9 & 48.2 & 60.3 \\
Gemma-3-4B
& 52.1 & 49.3 & 51.8 & 41.1 & 42.7 & 52.0 \\
Llama-3.1-8B
& 42.3 & 39.9 & 42.4 & 30.0 & 33.8 & 43.0 \\
\bottomrule
\end{tabular}

\caption{\textbf{Mean normalized judge scores of eight response generators on MATH-500 and Arena-Hard across numerical scoring topologies.} GPT-5-mini is used as the judge for all evaluations. Each value denotes the average score assigned by the judge to a response generator after converting the raw scores under each topology to a common 0--100 scale. The evaluated topologies include Likert-5 (0 to 5), Likert-10 (0 to 10), Linear-100 (0 to 100), Shifted Range (0.74 to 1.48), Narrow Range (0.3 to 0.7), and Irregular Range II ($-2.35$ to 7.85). Although absolute judge scores vary across scoring topologies, the relative ordering of all eight response generators remains unchanged on both benchmarks.}
\label{tab:response-generator-scores}
\end{table*}

\begin{table*}[t]

\subsection{Average Score Shifts and Ranking Stability}
\label{appendix:response-generator-score-shifts}

\vspace{6pt}

\centering
\small
\setlength{\tabcolsep}{7pt}
\renewcommand{\arraystretch}{1.08}

\begin{tabular}{lrr}
\toprule
\textbf{Scoring Topology}
& \textbf{MATH-500}
& \textbf{Arena-Hard} \\
\midrule
Likert-10
& $-0.1$
& $-3.4$ \\

Linear-100
& $+0.3$
& $-0.7$ \\

Shifted Range
& $-2.7$
& $\mathbf{-9.8}$ \\

Narrow Range
& $-2.1$
& $\mathbf{-9.4}$ \\

Irregular Range II
& $+0.6$
& $+0.1$ \\
\bottomrule
\end{tabular}

\caption{\textbf{Average normalized score shifts relative to Likert-5.} Values show the mean score change relative to Likert-5 across the eight response generators in Table~\ref{tab:response-generator-scores}. Negative values indicate lower average judge scores than Likert-5. The Shifted and Narrow Ranges produce the largest downward shifts on Arena-Hard, reducing the mean score by 9.8 and 9.4 points, respectively. Despite these shifts, the complete response-generator ranking is preserved on both benchmarks, with Kendall's $\tau=1.00$ and no top-1 or pairwise ranking flips.}
\label{tab:response-generator-score-shifts}
\end{table*}

\end{document}